\documentclass[journal]{IEEEtran}
\ifCLASSINFOpdf
\else
\fi

\usepackage{xspace}
\usepackage{xcolor}
\usepackage{graphicx}

\usepackage{enumitem}
\usepackage{url}
\usepackage{amsmath}

\begin{document}
%
\title{Towards Artificial General Intelligence (AGI) in the Internet of Things (IoT): Opportunities and Challenges}
%
%
%

\author{Fei Dou$^1$, Jin Ye$^2$, Geng Yuan$^1$, Qin Lu$^2$, Wei Niu$^1$, Haijian Sun$^2$, Le Guan$^1$, Guoyu Lu$^2$, Gengchen Mai$^3$, Ninghao Liu$^1$, Jin Lu$^1$, Zhengliang Liu$^1$, Zihao Wu$^1$, Chenjiao Tan$^4$, Shaochen Xu$^1$, Xianqiao Wang$^5$, Guoming Li$^6$, Lilong Chai$^6$, Sheng Li$^7$, Jin Sun$^1$, Hongyue Sun$^5$, Yunli Shao$^5$, Changying Li$^4$, Tianming Liu$^1$, Wenzhan Song$^2$

\thanks{$^1$Fei Dou, Geng Yuan, Wei Niu, Le Guan, Ninghao Liu, Jin Lu, Zhengliang Liu, Zihao Wu, Shaochen Xu, Jin Sun, and Tianming Liu are with the School 
of Computing, University of Georgia, Athens, GA 30602, USA (e-mail: \{fei.dou, geng.yuan, wniu, leguan, ninghao.liu, jin.lu, zl8864, zihao.wu1, shaochen.xu25, jinsun, tliu\}@uga.edu).}

\thanks{$^2$Jin Ye, Qin Lu, Haijian Sun, Guoyu Lu, and Wenzhan Song are with the School of Electrical and Computer Engineering, University of Georgia, Athens, GA 30602, USA (e-mail: \{jin.ye, qin.lu, hsun, guoyu.lu, wsong\}@uga.edu)}

\thanks{$^3$Gengchen Mai is with the Department of Geography, University of Georgia, Athens, GA 30602, USA (e-mail: gengchen.mai25@uga.edu)}

\thanks{$^4$ Chenjiao Tan and Changying Li are with Agricultural and Biological Engineering Department, University of Florida, Gainesville, FL 32611, USA (email: \{c.tan, cli2\}@ufl.edu)}

\thanks{$^5$Xianqiao Wang, Hongyue Sun, and Yunli Shao are with the School of Environmental, Civil Agricultural and Mechanical Engineering, University of Georgia, Athens, GA 30602, USA (e-mail: \{xqwang, hongyuesun, yunli.shao\}@uga.edu)}

\thanks{$^6$Guoming Li and Lilong Chai are with the Department of Poultry Science, University of Georgia, Athens, GA 30602, USA (e-mail: \{gmli, lchai\}@uga.edu)}

\thanks{$^7$Sheng Li is with the the School of Data Science, University of Virginia, Charlottesville, VA 22903, USA (e-mail: vga8uf@virginia.edu)}

}
\maketitle

\begin{abstract}

Artificial General Intelligence (AGI), possessing the capacity to comprehend, learn, and execute tasks with human cognitive abilities, engenders significant anticipation and intrigue across scientific, commercial, and societal arenas. This fascination extends particularly to the Internet of Things (IoT), a landscape characterized by the interconnection of countless devices, sensors, and systems, collectively gathering and sharing data to enable intelligent decision-making and automation. This research embarks on an exploration of the opportunities and challenges towards achieving AGI in the context of the IoT. Specifically, it starts by outlining the fundamental principles of IoT and the critical role of Artificial Intelligence (AI) in IoT systems. Subsequently, it delves into AGI fundamentals, culminating in the formulation of a conceptual framework for AGI's seamless integration within IoT. The application spectrum for AGI-infused IoT is broad, encompassing domains ranging from smart grids, residential environments, manufacturing, and transportation to environmental monitoring, agriculture, healthcare, and education. However, adapting AGI to resource-constrained IoT settings necessitates dedicated research efforts. Furthermore, the paper addresses constraints imposed by limited computing resources, intricacies associated with large-scale IoT communication, as well as the critical concerns pertaining to security and privacy.
\end{abstract}



%
\IEEEpeerreviewmaketitle

\section{Introduction}
\subsection{IoT Concept}
The Internet of Things (IoT) constitutes a transformative technological paradigm that envisions a world where physical objects and digital systems harmoniously interact to enhance the quality of human life and streamline various processes \cite{lakhwani2020internet,nord2019internet,laghari2021review}. In a simplified conceptual framework, IoT can be represented as a basic formula~\cite{farhan2017survey}:

\begin{equation}
\begin{split}
 IoT =   Humans + Physical~Objects (comprising~devices, \\
     sensors, controllers, storage, etc.) + Internet
\end{split}
\end{equation}
This captures the essence of IoT's interconnected nature, where human interactions, the physical world's components, and the vast expanse of the internet converge.

\begin{figure}[htbp]
   \centering
\includegraphics[width=0.48\textwidth]{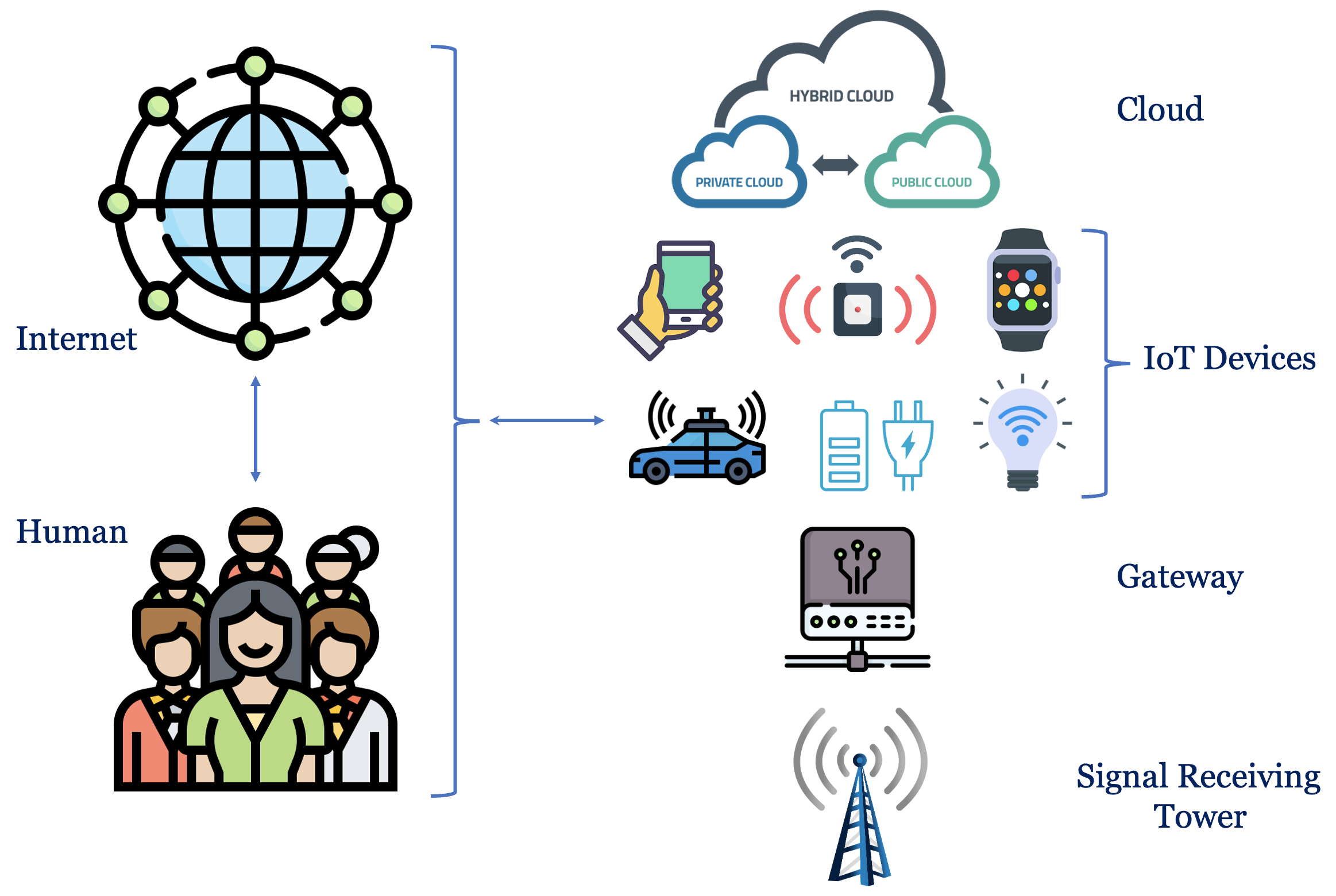}
   \caption{An illustration of IoT architectures.}
   \label{fig:iot}
\end{figure}

Fig. \ref{fig:iot} illustrates a fundamental depiction of the IoT system \cite{lohiya2020application, atzori2010internet}. Within this system, users and IoT devices are interconnected via the Internet. Communication between users and devices is facilitated by gateways and signals, which are transmitted and routed through signal receiving towers. Additionally, the data shared among these interconnected users and devices can be archived on the public cloud for later retrieval, while sensitive private data can be securely stored in the private cloud.

Based on the above paradigm, IoT devices have become omnipresent in everyday life, particularly due to the flourishing of AI techniques.
For example, connected appliances~\cite{bui2018consensual}, such as smart thermostats, refrigerators, and washing machines, are equipped with sensors that provide data about their status and performance.
Wearable devices with healthcare applications~\cite{kazanskiy2022recent,fang2016users}, such as Fitbit and Apple watch, are able to monitor various health metrics such as heart rate, steps taken, and sleep patterns, while synchronizing the data with smartphones or medical institutions, allowing users or doctors to track the physiological indicators in a real-time manner.
Smart cameras~\cite{pierce2019smart} are widely used for home security purposes, which are integrated with multiple sensors for motion detection, two-way audio communication, and facial recognition. 
The IoT devices are also used for connected vehicles~\cite{bui2019aco,mahmood2020connected} for features like GPS navigation, remote diagnostics, and even self-driving capabilities. These devices communicate with each other and external systems to enhance driver safety and convenience.

\subsection{AI in IoT} 

In recent times, artificial intelligence (AI) has exhibited remarkable achievements across various domains of IoT. AI's success has resonated strongly in IoT applications, powering tasks that encompass areas like sensor data analysis \cite{balakrishna2020iot, syafrudin2018performance}, predictive maintenance \cite{kanawaday2017machine, cheng2020data}, and real-time decision-making \cite{andronie2021artificial, syafrudin2018performance}. For instance, deep neural networks, including advanced architectures like Convolutional Neural Networks (CNNs) \cite{o2015introduction} and Recurrent Neural Networks (RNNs)~\cite{medsker2001recurrent}, have propelled IoT systems to autonomously process and interpret complex visual and textual data streams. These networks enable smart devices to excel in intricate tasks such as anomaly detection in manufacturing processes \cite{scime2018anomaly} and natural language understanding for voice-controlled home automation \cite{rani2017voice}. Additionally, AI models like RoBERTa \cite{liu2021robustly} have demonstrated their prowess in enhancing human-like comprehension of IoT-generated textual data. It is important to note that while many AI advancements have traditionally focused on surpassing human performance in single cognitive abilities, the convergence of deep learning and IoT ushers in a new era where foundational models, pre-trained on expansive multimodal datasets, can be swiftly adapted to a diverse array of downstream cognitive tasks, marking a significant stride towards achieving Artificial General Intelligence (AGI) in the context of IoT.

\subsection{AGI Background}

The fast development of AI in the past ten years can be attributed to three major reasons: the increasing availability of training data, enhanced support from AI infrastructures, and the advancement of AI models such as Generative Adversarial Networks (GAN)~\cite{goodfellow2020gan}, diffusion models~\cite{ho2020ddpm}, and Transformers~\cite{vaswani2017attention}.

Recently, AGI has become an increasingly popular topic, not only within the realm of AI but also across diverse fields such as agriculture \cite{lu2023agi,rezayi2023exploring}, biology \cite{ma2023towards}, healthcare \cite{Lehman2023do}, geography \cite{mai2023csp,mai2023opportunities}, and so on, especially after OpenAI's recent introduction of ChatGPT~\cite{openai2022chatgpt}. 
In fact, we believe the technological advancement of AGI should not be linked solely to a particular entity; rather, it embodies a collaborative endeavor between academia and industry over recent years. The triumph of AGI, notably concerning large language models (LLMs), finds its origins in the introduction of the Transformer architecture~\cite{vaswani2017attention} by Google in 2017. This innovative architecture, which revolves around parallelizable self-attention based sequential neural networks, was conceived as a substitute for the RNN.
Subsequently, a milestone in the history of the LLMs and AGI is the development of BERT (Bidirectional Encoder Representations from Transformers)~\cite{kenton2019bert}. Through the process of pre-training on extensive unsupervised text data followed by supervised fine-tuning on labeled data, BERT accomplished groundbreaking results across eleven natural language processing (NLP) tasks.

The success of BERT started a new era of AI, i.e., AGI -- instead of training task-specific AI models, an increasing number of researchers started to seek ways to first pre-train a large model on internet-scale data in a task-agnostic manner and then adapt this model to various downstream tasks via fine-tuning, few-shot learning, or even zero-shot learning. This genre of large-scale task-agnostic models, later on, is called \textit{foundation models}~\cite{bommasani2021opportunities,wei2022palm,ouyang2022instructgpt,openai2022chatgpt,mai2022towards,kirillov2023sam,mai2022towards}. Large language models are examples of foundation models: GPT-3~\cite{brown2020gpt3}, InstructGPT~\cite{ouyang2022instructgpt}, ChatGPT~\cite{openai2022chatgpt}, GLaM \cite{du2022glam}, PaLM \cite{chowdhery2022palm}, LLaMA~\cite{touvron2023llama}, Alpaca \cite{taori2023alpaca}, Claude2 \footnote{\url{https://www.anthropic.com/index/claude-2}}, LLaMA 2 \cite{touvron2023llama2}, and so on. 

The development of LLMs in the NLP domain also inspired the development of foundation models in other domains such as reinforcement learning (RL)~\cite{reed2022gato} and computer vision (CV)~\cite{rombach2022stablediffusion,kirillov2023sam}. For example, many diffusion-based models such as DALL$\cdot$E$\cdot$2~\cite{ramesh2022dalle2}, Imagen~\cite{saharia2022imagen}, Stable Diffusion~\cite{rombach2022stablediffusion} are visual foundation models. They were trained in a task-agnostic manner but can be adapted to many vision tasks such as style transfer~\cite{ramesh2022dalle2}, image editing~\cite{meng2021sdedit,saharia2022imagen}, image denoising~\cite{saharia2022palette,kawar2022ddrm}, image super-resolution~\cite{saharia2022sr3}, image inpainting~\cite{saharia2022palette,ramesh2022dalle2,rombach2022stablediffusion}, image colorization~\cite{saharia2022palette,kawar2022ddrm,rombach2022stablediffusion}, image compression~\cite{kawar2022ddrm}, and so on. Most of the current diffusion-based visual foundation models focus on the image-to-image translation problem. 
Recently, the Segment Anything Model (SAM)~\cite{kirillov2023sam} has been proposed as a visual foundation model for various segmentation tasks in remote sensing~\cite{zhang2023text2seg} and the medical~\cite{mazurowski2023sammedical} domain.

In addition, instead of limiting foundation models to one data modality, one rising trend in foundation model research is developing multimodal foundation models which can simultaneously handle various data modalities such as text, images, video, audio, and so on. An early pioneering work is CLIP \cite{radford2021clip} which pre-trained a text encoder and an image encoder jointly with a self-supervised contrastive learning objective based on a paired text-image dataset. The benefits include knowledge sharing across data modalities and the ability to enable tasks that require multiple data modalities. This practice inspired many follow-up studies on multimodal foundation models such as BriVL \cite{fei2022BriVL}, BLIP \cite{li2022blip}, GPT-4 \cite{openai2023gpt4}, KOSMOS-1 \cite{huang2023kosmos1}, and KOSMOS-2 \cite{peng2023kosmos2}.

\begin{figure*}[htbp]
   \centering
\includegraphics[width=0.72\textwidth]{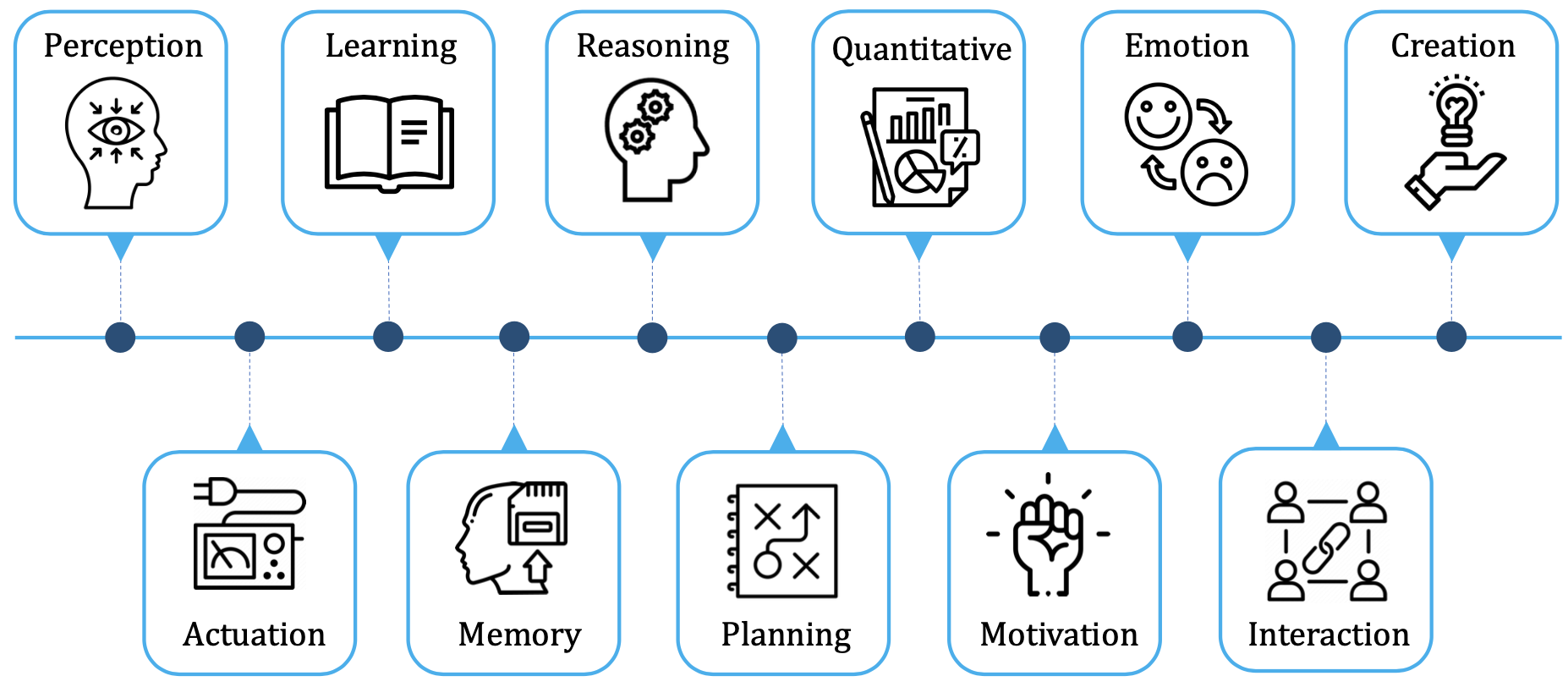}
   \caption{A summary of AGI competencies.}
   \label{fig:agi_com}
\end{figure*}

\subsection{Towards AGI Paradigm for IoT}
A point broadly shared in the AGI community is called the core AGI hypothesis~\cite{goertzel2014artificial}, which refers to ``the creation and study of synthetic intelligence with sufficiently broad scope and strong generalization capability, is qualitatively different from the synthetic intelligence with significantly narrower scope and weaker generalization capability".
We characterize the AGI systems with a list of capabilities motivated by the competencies summarized by AI researchers and psychologists~\cite{adams2012mapping} in Fig. \ref{fig:agi_com}.
\begin{enumerate}[itemsep=0pt]
    \item \textbf{Perception}: The sense and understanding of visual, hearing, touch information from environment, and the ability of integrating multi-modal information from various senses.
    \item \textbf{Actuation}: The ability of manipulating physical objects, using tools, and navigation in complex environments.
    \item \textbf{Memory}: The memory regarding facts or beliefs, the outcome of sequential/parallel combinations of actions, and experience attributed to a particular instance.
    \item \textbf{Learning}: The ability to learn from teachers, other observed agents, media, experimentation, and positive or negative reinforcement signals in the environment.
    \item \textbf{Reasoning}: The ability of deduction, induction, and abduction. The ability of reasoning from observed premises, physical rules and spatio-temporal associations.
    \item \textbf{Planning}: The ability of conducting strategical, physical, and social planning.
    \item \textbf{Motivation}: The ability of creating sub-goals based on the pre-programmed goals, or driven by curiosity, emotions, empathy, and altruism.
    \item \textbf{Emotion}: The ability of expressing emotion, as well as perceiving or interpreting emotion.
    \item \textbf{Interaction}: The ability to initiate communication and organize group activities, with appropriate behavior. The communication could be achieved through verbal, gestural, pictorial or even cross-modal signals.
    \item \textbf{Quantitative}: The ability to comprehend and articulate mathematical concepts, solve mathematical problems, and apply quantitative reasoning to solve problems that demand mathematical thinking and model-building skills.
    \item \textbf{Creation}: The ability to build and modify physical objects, assemble and organize social groups, and form novel concepts.
\end{enumerate}
AGI within the IoT context would imply the development of highly advanced and versatile AI systems that possess human-like cognitive capabilities (i.e., more than one capability listed above) and can seamlessly interact and adapt across a broad spectrum of IoT devices and scenarios.

\subsection{AI in IoT Applications and the Necessity for AGI}
IoT generates massive amounts of heterogeneous data from interconnected sensors, devices, and systems \cite{marjani2017big}. Applying AI to analyze and extract value from this data is critical. Some common AI use cases in IoT include:

\begin{enumerate}[itemsep=0pt]
    \item \textbf{Predictive Maintenance:} Using machine learning to analyze sensor data from industrial equipment and predict failures before they occur \cite{durbhaka2021convergence}. This avoids costly IoT downtime.
    \item \textbf{Anomaly Detection:} Identifying anomalies in real-time IoT data streams to detect cyberattacks, equipment faults, or other issues requiring intervention \cite{sgueglia2022systematic}. 
    \item \textbf{Personalization:} Applying deep learning to user data from smart devices to provide personalized recommendations and customized experiences \cite{dhelim2021iot}.
    \item \textbf{Smart Grids:} AI algorithms analyze data from smart meters and sensors to optimize energy distribution, manage demand response, and forecast energy needs, enhancing the reliability and efficiency of power grids \cite{ali2020state}.
    \item \textbf{Smart Home:} AI-driven systems in smart homes \cite{murdan2023smart} learn user preferences and behavior to automate lighting, heating, and cooling, resulting in an intelligent environment that enhances comfort and reduces energy consumption.
    \item \textbf{Smart Agriculture:} AI enhances smart agriculture by processing data from sensors and satellite images for precision farming, optimizing irrigation, and predicting environmental impacts on crop yields \cite{yang2022deep, yang2023sam, shaikh2022machine}.
    \item \textbf{Smart Robotics:} AI-driven robotic systems achieve self-perception and environmental perception in robots and drones by collecting and analyzing real-time, multimodal data from sensors and cameras. This enables precise navigation and manipulation \cite{liu2023digital} across a wide range of applications, from last-mile delivery to industrial automation and automated farming.
    \item \textbf{Smart Manufacturing:} In industrial IoT, AI is used for modeling, monitoring, diagnosis, and control by analyzing machine data to forecast breakdowns, optimizing production schedules, and improving supply chain efficiencies \cite{soori2023internet}.
    \item \textbf{Smart Healthcare:} AI applications in smart healthcare \cite{al2022iot} include predictive analytics for patient monitoring, natural language processing for electronic health records \cite{li2023artificial,holmes2023evaluating,liu2022survey}, and image analysis for diagnostic imaging \cite{li2023artificial,liu2023summary}.
    \item \textbf{Smart Transportation:} AI analyzes data from sensors and cameras to optimize traffic flows, improve public transportation systems, enhance navigation with real-time data, and facilitate autonomous vehicles \cite{bharadiya2023artificial}.
    \item \textbf{Smart Public Safety:} AI helps in analyzing data from IoT devices for faster emergency response, uses pattern recognition to anticipate incidents, and improves communication systems in crisis situations \cite{sun2020applications}.
    \item \textbf{Smart Environmental:} AI is utilized in smart environmental systems to analyze data from sensors for pollution control, predict environmental trends, and facilitate wildlife monitoring for conservation efforts \cite{el2013smart}.
\end{enumerate}

While traditional AI applications perform well for narrow tasks, developing more versatile systems requires AGI. Conventional AI lacks the flexible reasoning, contextual adaptation, and learning capabilities of human intelligence \cite{zhao2023brain}. AGI aims to produce AI with greater generalizability across tasks and situations. Key capabilities for AGI include:

\begin{enumerate}[itemsep=0pt]
    \item \textbf{Transfer learning:} Applying knowledge gained in one domain to new domains \cite{zhou2023comprehensive}.
    \item \textbf{Multimodal understanding:} Integrating and contextualizing data from diverse sensors (text, audio, video, etc.) \cite{li2023artificial,zhou2023comprehensive}.
    \item \textbf{Reasoning:} Understanding causes and effects to explain events and make predictions \cite{wei2022chain}.
    \item \textbf{Self-supervised learning:} Discovering higher-level concepts and representations by exploring environments \cite{zhao2023brain,liu2023summary}.
    \item \textbf{Interactivity:} Modern generative AGI models are capable of dynamic and interactive engagements with humans and environmental inputs, bringing us closer to the goal of truly interactive agents \cite{liu2023summary}.
    \item \textbf{Few-shot learning:} Learning new skills and apply to unseen tasks rapidly from limited examples \cite{brown2020language}.
\end{enumerate}

AGI will enable the creation of AI agents that can dynamically process and analyze multifaceted IoT data, understand context, and make decisions accordingly. This will unlock more powerful, flexible applications rather than current narrow AI solutions. 

\subsection{Challenges}
IoT devices are destined to be the main carrier for future AI applications.
However, the confluence of AGI and IoT devices is still facing many challenges.

AGI models, such as the LLMs, exhibit immense computational demands, while IoT devices are often characterized by their limited computing power (e.g., computation resource and memory size).
This is a natural challenge when accommodating resource-intensive AGI.
IoT devices also have a limited energy budget since most of them are designed to be battery-powered. However, intensive computation of AI models could drain the battery quickly.

Other than executing the computation of AI models locally, the computation can also be offloaded to the centralized data center. In this paradigm, IoT devices are mainly responsible for gathering raw data from the physical environment via the embedded sensors and transmitting this data to central repositories or other connected devices.
However, the bandwidth limitations and communication in the dynamic environment of IoT networks~\cite{dai2020efficient} can hinder the timely transfer of data to centralized AGI processing nodes.

The integration of AGI with IoT devices also magnifies challenges surrounding data privacy and security. 
The distributed nature of IoT devices and the data-hungry nature of AGI amplify vulnerabilities, necessitating robust measures for authentication, encryption, and access control. 
IoT devices gather diverse and often sensitive data, which are required by AI models for training and inference. 
However, deploying AI models on IoT devices that respect data privacy presents a formidable challenge. 
Models trained on personal or sensitive data, e.g., the LLMs fine-tuned with users' personal behavior patterns or habits, may inadvertently expose confidential information during inference, raising privacy, ethical, and legal issues.

Another challenge lies in ensuring the lifespan and adaptability of AGI over time. IoT devices often find their way into infrastructure and devices with extended lifespans.
Though AGI could achieve much better generalizability compared to the conventional AI application that mainly targets solving specific problems, the integration of AGI within IoT devices necessitates foresight into or adapting to the evolving deployment scenarios over time.

\section{AI Foundation for IoT}





\subsection{IoT-Centric Foundation Model for Achieving AGI}
As mentioned previously, the emergence of IoT has led to a broad spectrum of applications across various domains, including power grids, healthcare, and smart cities, among others. 
Despite their diverse forms, these IoT-related applications can be mathematically abstracted as a function $f(\cdot)$ that establishes a mapping between input features ${\bf x}$ and the resulting output $y$. With ${\bf x}$ representing the measured vitals for wearables, the output $y$ could be a binary indicator for heart attack in smart healthcare. 
Building on data modality, the prevalent approaches for IoT data analysis can be broadly categorized into CNNs for images, RNNs for times series, and graph neural networks (GNNs) for networked data. 
Albeit achieving encouraging performances, the aforementioned IoT-centered approaches remain tailored to specific tasks and domains. For each problem at hand, the sought learning model has to be retrained from scratch. Apparently, this paradigm is far away from biological intelligence that is capable of transferring knowledge across domains and tasks. 
Moving towards AGI, our vision is to leverage the \textit{foundation models} (FMs), empowered by vast quantities of training data, learning parameters, and computational resources~\cite{bommasani2021opportunities}. The application of FMs emerges in the field of NLP, and are later adapted to the computer vision community. Most recently, with the advent of models like ChatGPT, FMs have garnered significant attention and have been successfully applied to diverse domains such as education, geoscience, and agriculture. However, in the realm of IoT data analysis, the potential of FMs remains largely unexplored.

\subsubsection{Basic components of FM-based learning}

Given a pre-specified FM that is equipped with massive number of parameters, one starts with pre-training the model using a huge number of data samples, and subsequently adapted to domain-specific tasks via fine-tuning techniques. Next, we will outline the three basic modules involved in the FM-based AGI paradigm, that is, Transformer-based model structure, pre-training, and adaptation.

\textbf{Transformer}
The Transformer is undoubtedly the most celebrated architecture underlying most FMs. Unlike CNNs and RNNs, it solely relies on the attention mechanism to allow for the transfer of weighted representation knowledge between various neural units~\cite{vaswani2017attention}.  Specifically, it assigns weights to all the encoded input representations and learns the most important part of the input data. Numerous attention mechanisms have been developed in large models~\cite{guo2022attention}. Today, attention-based Transformer is the most popular structure for FMs in NLP and CV. In NLP, the Transformer can characterize the long-range dependency in the sequential input data. For example, the GPT-3~\cite{brown2020language} is a generative model based on the transformer. The Vision Transformer (ViT)~\cite{dosovitskiy2020image} in CV is proposed to represent an image to a series of image patches. The number of parameters in the aforementioned Transformer-based FMs is large; e.g.,  $175$ and $22$ billion for GPT-3 and ViT-22B~\cite{dehghani2023scaling}, respectively. In spite of these large scales, Transformers can still be scalable thanks to the model parallelization.

\textbf{Pre-training}
Given enormous (unlabelled) data, the FM is firstly trained on {\it pre-training tasks} to obtain specific attributes, structure, and feature representations that could later on 
be adapted to downstream tasks for faster convergence. These pre-set tasks are tailored to the data modality and application domain, and are usually learned in a self-supervised fashion, which extracts important feature mappings without labelled data. Generally, self-supervised learning (SSL) approaches can be classified into generative- and discriminative-based. For the former, variational autoencoder (VAE) ~\cite{kingma2013auto} and generative adversarial network (GAN)~\cite{goodfellow2020generative} are two representative examples, which are to reconstruct the data itself. For the latter, contrastive learning is a widely adopted discriminative SSL method in CV and NLP. The main idea of contrastive learning is to learn a model that makes similar instances closer in the projected space, and dissimilar instances farther apart in the projected space. 

\textbf{Adaptation}
Given a pre-trained FM, application to a specific task entails model adaptation by incorporating new information. This can be achieved through task specification as in text summarization by appending a prompt to the input article, or alternatively by fine-tuning  the FM parameters using domain-specific data~\cite{bommasani2021opportunities}.


\subsubsection{FM-based AGI for IoT data}
The sheer volume of IoT data makes large FM-based AGI possible. Targeting at IoT-oriented FM learning, we identify two unique features of IoT data, namely: temporal evolution and multi-modality. The former entails fast temporal adaptation with low latency, the latter necessitates sophisticated technique to incorporate multiple sources of data.
In the subsequent section, we will delve into the strategies for tackling these two challenges, drawing insights from established methodologies within the realms of NLP and CV.

\textbf{Continual adaptation with low latency.}
Data collected by IoT sensors and devices usually vary unpredictably over time. This temporal dynamics induces distribution shift, thus rendering pre-trained FMs inaccurate without carefully designed adaptation techniques. Temporal adaptation for FMs has been investigated in the context of language modelling by reweighting the training data~\cite{lazaridou2021pitfalls}, explicitly conditioning a language model on the time period~\cite{dhingra2022time}, as well as retrieval based approaches (e.g., ~\cite{lewis2020retrieval}).


While the aforementioned adaptation schemes can be transferred to the IoT-related tasks, one has to take into account the time constraints. How to design adaptation methods with low latency and high accuracy is a major challenge facing FM-based AGI.research for IoT.
In addition to the adaptation step, innovative FM architectures and pre-training tasks are called for to better cope with such unknown temporal dynamics in IoT data.

\textbf{Accounting for multi-modalities}
IoT-based inference tasks usually rely on collected data from multiple sources. In smart medical diagnosis for example, one could rely on measurements from multiple wearable devices . How to cope with such multi-modaility poses another challenge for FM-based AGI in IoT. Drawing inspirations from vision-language multimodal FMs, we envision two ways to account for the interaction among different modalities according when the information fusion acorss multimodalites is conducted. While the first approach combines extracted feature representations from different modalities as in CLIP~\cite{radford2021clip}, the second method alternatively obtains feature vector after fusing the original multimodal data similar to ViLBERT~\cite{cho2021unifying}. Going beyond the aforementioned two frameworks, more sophisticated mechanisms are to be developed toward efficiently and effectively extracting and combining information from the multiple modalities. More sophisticated mechanisms are to be developed toward efficiently and effectively extracting and combining information from the multiple modalities.   


\subsection{Domain Generalization Techniques}
Domain generalization (DG) research has undergone a rapid acceleration, resulting in an abundant array of principled algorithms that can be broadly classified into four categories: domain alignment, meta-learning, data augmentation, and distributionally robust optimization~\cite{zhou2022domain, wang2022generalizing}. 

\textbf{Domain alignment.}
Domain alignment is the most popular category in domain generalization, which aims to minimize differences among different domains and learn domain-invariant representations~\cite{muller2022learning,zhou2022domain,peters2016causal,pearl2009causality}. Predictors that rely on the causes of the label to make predictions are created as a result. Several methods, such as invariant risk minimization~\cite{arjovsky2019invariant} and related ones~\cite{ahuja2020invariant, pezeshki2021gradient, krueger2021out, robey2021model, zhang2021can}, have been proposed based on the invariance principle from causality. This principle distinguishes predictors that solely rely on the causes of the label from those that do not. The optimal predictor that only focuses on the causes is invariant and min-max optimal~\cite{rojas2018invariant, koyama2020out, ahuja2021invariance} under distribution shifts, but the same is not true for other predictors.

To reduce distribution mismatch in domain generalization, one approach is to learn a mapping function that can minimize the moments, which is a representation of the distribution, of the transformed features between source domains~\cite{muandet2013domain, erfani2016robust, li2018domain, jin2020feature}. Another option for reducing distribution mismatch is to take the semantic labels into account~\cite{motiian2017unified, mahajan2021domain}. The basic idea is to construct the anchor group, the positive group (same class as the anchor but from different domains), and the negative group (different class than the anchor). To achieve this, the anchor and the positive groups are pulled together, while the anchor and the negative groups are pushed away~\cite{motiian2017unified, mahajan2021domain}.
Commonly used distribution divergence measures between two probability distributions are also applied to align the domains~\cite{li2018domain, wang2021respecting, 8058000}. In domain generalization, adversarial learning~\cite{creswell2018generative} is also performed between source domains to learn source domain-agnostic features that are expected to work in novel domains. Simply speaking, the learning objective is to make features confuse a domain discriminator, which can be implemented as a multi-class domain discriminator~\cite{9806715, 8909793, 8758199, li2018deep, rahman2020correlation, deng2020representation, aslani2020scanner}.

\textbf{Meta-learning}
Meta-learning, also known as learning-to-learn, is a rapidly developing field that has found applications in domain generalization. The main idea behind using meta-learning for DG is to train a model on a range of tasks that involve domain shift, with the aim of improving its performance on new tasks with unseen domains. For instance, MAML~\cite{finn2017model} trains a model on meta-train and meta-test sets to enhance its performance on the meta-test set. Many studies~\cite{zhao2021learning, du2021metanorm, dou2019domain, li2018learning, li2019feature, li2021sequential, liu2020shape} have focused on two crucial components in meta-learning, namely episodes, which are formed from available samples, and meta-representation, which is defined in~\cite{hospedales2021meta} to represent the model parameters that are meta-learned.

\textbf{Data augmentation}
To enhance model performance in testing domains, data augmentation is a natural approach that involves creating new $(A(x), y)$ pairs by applying transformations to original $(x, y)$ pairs.
Based on the type of transformation used, existing methods can be classified into four categories:
a) hand-engineered image transformations~\cite{volpi2019addressing, zhang2020generalizing, shi2020towards}. 
b) adversarial gradients obtained from category or domain classifiers~\cite{volpi2018generalizing, zhou2020deep, qiao2020learning, zhao2020maximum,zhu2022crossmatch}. 
c) model $A(\cdot)$ using neural networks, such as random CNNs~\cite{xu2020robust}, an off-the-shelf style transfer model~\cite{zhou2021semi, borlino2021rethinking}, or a learnable image generator~\cite{zhou2020learning}. 
d) inject perturbation into intermediate features in the task model~\cite{zhou2021domain, zhou2021mixstyle}.

\textbf{Domain adaptation}
Distributionally robust optimization (DRO), a method to optimize for worst-case loss over potential test distributions, is also a useful approach to avoid learning spurious correlations that hold on average but not in atypical groups~\cite{ben2013robust, duchi2021statistics, oren2019distributionally, shafieezadeh2015distributionally, namkoong2017variance, duchi2018learning,hajifar2022online}.

The paper\cite{khan2018scaling} presented an unsupervised domain adaptation method that also incorporates target domain data during training. Their approach involves minimizing the KL divergence between the representations of the source and target domains for each layer's output in the model. This objective aims to eliminate the discrepancy between the source and target domains.  
\cite{wang2018deep} introduced a source domain selection algorithm aimed at identifying the most similar source domains by comparing the cosine similarities of each source domain with the target domain. Additionally, they minimized the distance between representations in the last fully connected layers of both the source and target domains.
On the other hand, \cite{ganin2016domain} proposed a gradient reversal layer on the feature extractor. This layer's purpose is to reverse the gradient direction during the back-propagation stage, enabling adversarial training to encourage the model to learn invariant features across domains. \cite{wilson2020multi} adopted the same strategy as \cite{ganin2016domain} in their work.
\cite{mazankiewicz2020incremental} keep the batch norm layer active during inference stage which could adapt to online mode as well.

According to \cite{richard2021unsupervised}, they claim to be the first ones in applying adversarial unsupervised domain adaptation to regression tasks. Their model comprises two prediction layers, denoted as $h1$ and $h2$. The algorithm in \cite{richard2021unsupervised} involves four optimization steps during training:
(1)Minimizing the prediction loss of $h1$ on source domain. (2) Maximizing the discrepancy between the source and target domains by updating parameters in $h2$.




\section{Applications of IoT towards AGI: Enhancements and Potential Opportunities}

In this section, we provide an overview of the primary application scenarios of IoT and their integration with the AGI paradigm.

\begin{figure*}[htbp]
   \centering
    \includegraphics[width=0.97\textwidth]{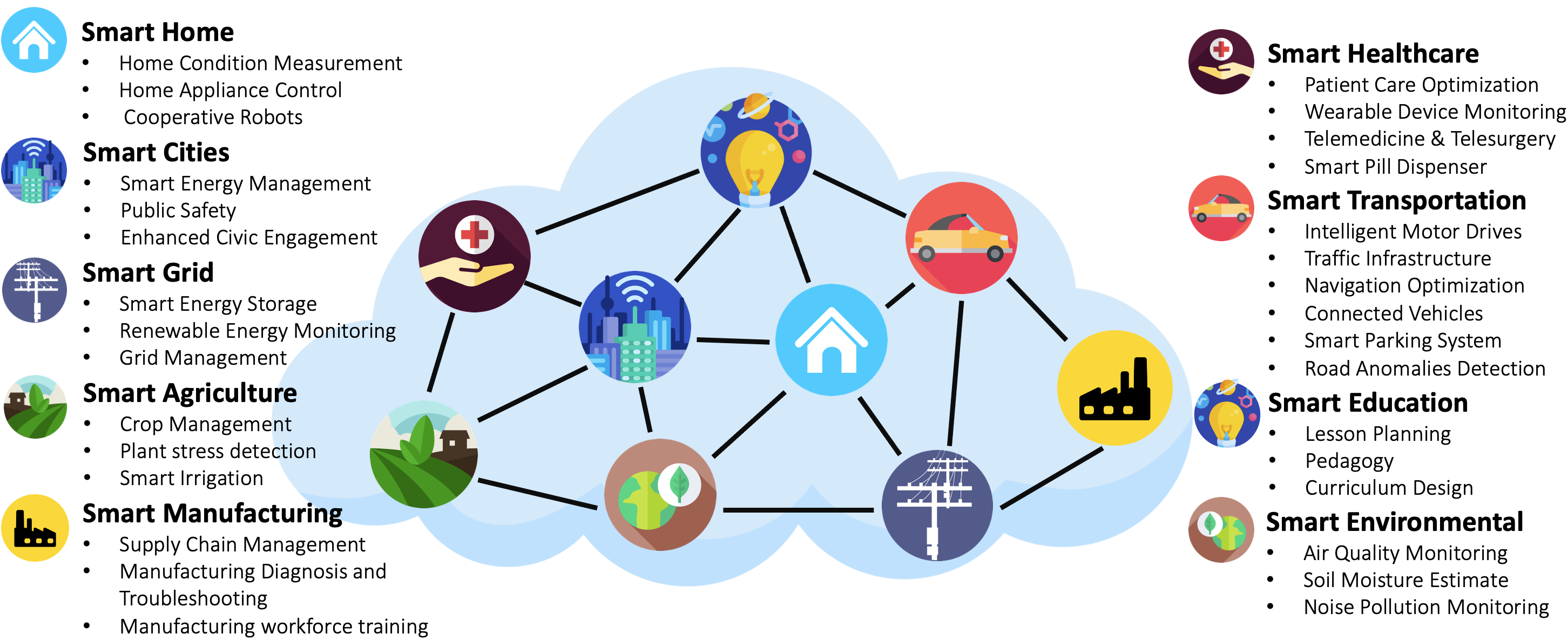}
   \caption{Examples of IoT applications.}
   \label{fig:iot_app}
\end{figure*}
\subsection{Smart Grid}
Over the past decade, concerns about cybersecurity in intelligent power electronics systems have grown due to the widespread implementation of networked digital control units. Numerous studies~\cite{ye2022review,zhang2021physical,yang2020vulnerbility,ye2021cyber} have demonstrated the vulnerabilities and impacts of modern power electronics systems across various applications, such as photovoltaic (PV) systems, electric vehicles, and intelligent manufacturing systems. Recent research has focused on different detection approaches targeting diverse power electronics applications, including DC microgrids, PV farms, and industrial motor drives, to address these concerns and enhance the reliability of intelligent power electronics systems. For IoT enabled power electronics systems, we focus on two sub-applications: power-electronics-based smart grid and intelligent motor drives.

Power electronics converters play a crucial role in integrating distributed energy resources (DERs) with the power grid. The increasing use of industrial Internet of Things (IoT) edge devices has led to the incorporation of various built-in functionalities in power converters, such as remote control and wireless communication with a central plant controller. Recent literature highlights the use of such functionalities, including converter remote control and wireless communication, to enhance power converter operation ~\cite{lasagani2018communication,terashmila2017comparison,yang2011communication}. However, the latest revisions to IEEE 1547 standard mandate a set of control parameters for grid-tied DER converters that require remote control and adjustment by a Supervisory Control and Data Acquisition (SCADA) system via a communication network ~\cite{sahoo2019cyber,roslan2018remote,kim2008dynamic}. With the increasing number of DER equipment connected to the IoT infrastructure, power electronics based smart grids are more susceptible to cyber attacks.

Detection of cyber-attacks in smart grids has become an essential topic because more internet technologies are employed in communication between distributed energy resources and central locations~\cite{lasagani2018communication,terashmila2017comparison}. Smart grid cyber-physical security studies have proposed many cyber-attack detection and diagnosis methods using AI techniques. In~\cite{8963972}, RNN is used for false data injection attack detection in a DC microgrid with dynamic loads. In~\cite{8526328}, a novel cooperative mechanism based on a secondary voltage controller is proposed to facilitate the detection and mitigation of stealth attacks on DC microgrid. In~\cite{9060989}, a measurement data authentication algorithm based on Fast Fourier Transform (FFT) and machine learning methods is developed for smart grids to protect against data spoofing attacks. 
 In~\cite{haghnegahdar2020}, a novel optimization technique is developed to train an Artificial Neural Network (ANN) to classify and detect cyber-attacks and intrusions in a smart grid. In~\cite{habibi2020detection}, an artificial intelligence-based method is proposed for detection of DIAs in DC microgrids. By exploiting the nonlinear mapping capability of nonlinear autoregressive exogenous (NARX) neural networks, cyber-attacks in DC microgrids could be identified. Although a supervised learning method could generate a good estimation or classification result to distinguish cyber-physical threats, it needs a large amount of data from the plant, either simulated or actual, during the training phase. However, it is impossible to emulate all of the potential cyber-physical threats in real applications. Specifically for a PV system,~\cite{li2020detection} proposed a multilayer LSTM-based diagnosis solution for DIAs in a two-stage PV converter. By using the collected electrical waveform, a supervised-learning-based classifier is trained for cyber-attack detection. In~\cite{li2020data}, an attempt is made to detect typical cyber-attacks in PV systems by using micro phasor measurement units ($\mu$PMU) data. Notice that the above two studies do not consider physical faults because electric waveforms and $\mu$PMU under cyber-attacks and physical faults show very similar patterns. To overcome the drawback of supervised learning methods,~\cite{li2019detection} proposed a binary matrix factorization-based cyber-attack diagnosis without a training process. The results have shown that most cyber-attacks are clustered at different locations.  Physics-informed neural networks  offer a promising avenue for addressing these concerns. This is achieved by seamlessly integrating principles rooted in physics (referred to as physics-informed) into the cutting-edge framework of deep learning. These paradigms encompass diverse aspects, including the formulation of physics-informed loss functions, the initialization of neural networks with physics-driven insights, the architectural design guided by physical principles, and the amalgamation of physics-deriven knowledge with deep learning models in a hybrid fashion  \cite{ huang2022applications}. In \cite{guo2021data},  innovative physics-guided features such as frequency-domain magnitude-based  residuals time-domain mean current vector-based feature were proposed to address novel cyber-attacks that are excluded from the machine learning training process.

\subsection{Smart Homes}
Advancements in electronics, information communication technologies, mobile applications, autonomous systems, virtualization, and cloud computing have contributed to the evolution of the smart home concept. 
A smart home is characterized as a living space outfitted with devices that possess computational capabilities and communication technologies. This setup seamlessly integrates diverse IoT devices and sensors, AI algorithms, and network connectivity. The purpose is to facilitate smooth communication and control of various systems and appliances within the household. 
Through collaborative interaction, these IoT devices ensure convenience, comfort, security, energy efficiency, and an improved quality of life for residents. Smart homes offer remote user control, enabling a wide range of tasks, such as voice interaction, setting alarms, managing to-do lists, and controlling other smart devices like locks, light bulbs, thermostats, and more~\cite{hammi2022survey, hui2017major, balakrishnan2018smart}. 

A smart home is characterized by its automation capabilities~\cite{stolojescu2021iot, jabbar2019design,singh2018iot,domb2019smart}, allowing tasks such as adjusting lighting, temperature, and blinds to be executed based on occupants' preferences or environmental conditions. The interconnected IoT devices communicate through a centralized hub or cloud-based platform, while remote access and control enable homeowners to manage devices from afar using smartphone apps or web interfaces~\cite{samuel2016review, pradeep2016iot,liu2020constructing, kumar2014android}. Energy efficiency~\cite{vishwakarma2019smart, reinisch2011thinkhome,ayan2020iot,hossain2017cyber} is promoted through smart thermostats, lighting, and energy monitoring systems to reduce consumption and utility bills. Enhanced security and surveillance features~\cite{pierce2019smart, yang2019security, kaushik2022isecurehome, dwivedi2021security}, like cameras, motion sensors, and smart locks, bolster safety and real-time monitoring. Voice-activated virtual assistants~\cite{yue2017voice, canziani2021consumer, ruslan2021development}, such as Amazon Alexa or Google Assistant, facilitate device control via voice commands. Smart homes also adapt to occupants' preferences, delivering personalized experiences~\cite{dahlgren2021personalization,bianchi2019iot,xie2019personalized,gourav2021personalization}and automating routines. Moreover, certain smart homes incorporate health monitoring devices for tracking vital signs, activity levels, and sleep patterns to promote wellness and safety for residents~\cite{umair2021impact,mohammed2020internet,talal2019smart,maswadi2020systematic}.

Large language models built upon the Transformer architecture, such as BERT \cite{kenton2019bert}, GPT \cite{radford2018improving}, XLNet \cite{yang2019xlnet}, and T5 \cite{raffel2020exploring} have introduce both opportunities and challenges towards AGI-integrated smart home. To address challenges including the models' dependence on discrete tokens for input, or  domain mismatch
problems \cite{mehrish2023review}, dedicated frameworks have been developed based on large language models. Advances in technologies like XLS-R \cite{babu2021xls}, Whisper \cite{radford2023robust}, and DALL-E \cite{wang2023neural} contribute to the advancement of AGI in smart homes by enhancing the capabilities of AI systems in understanding, communicating, and adapting to the dynamic and complex environments of modern living spaces. 

XLS-R is a large-scale model for cross-lingual speech representation learning based on wav2vec 2.0 \cite{baevski2020wav2vec, xu2021self}, where cross-lingual transfer is employed to enhance representations for low-resource languages using the knowledge from high-resource languages. XLS-R's ability to understand and process multiple languages enables AGI systems to comprehend diverse user commands and queries, catering to a global user base in smart homes. Understanding various languages helps AGI systems recognize cultural nuances, which can be crucial for personalized and context-aware interactions in smart homes.

The Whisper model \cite{radford2023robust} has revolutionized speech recognition, delivering exceptional accuracy across diverse speech tasks and challenging conditions. Through minimal data pre-processing and weak supervision, it achieves state-of-the-art results, excelling in multilingual recognition, translation, and language identification. Its accuracy in understanding spoken language can enable more intuitive and natural interactions between users and their smart home systems. This includes controlling devices, requesting information, and even having context-aware conversations. This level of communication bridges the gap between human language and technology, bringing smart home interactions closer to human-like interactions, a significant step towards AGI.

Recent strides in text-to-speech synthesis have led to innovative models like VALL-E \cite{wang2023neural}. This pioneering approach capitalizes on abundant semi-supervised data to train a versatile text-to-speech system, capable of producing personalized, high-quality speech. It offers diverse outputs while maintaining the acoustic setting and speaker's emotions related to the prompt. Incorporating VALL-E into AGI systems for smart homes elevates the quality of communication, personalization, and adaptability. These advancements bring us closer to the vision of AGI-powered smart homes that seamlessly interact with users in ways that are natural, tailored, and responsive, fostering an environment that aligns with users' preferences and needs.






\subsection{Smart Healthcare}
Incorporating AGI into the healthcare sector presents a vast potential for optimizing patient care \cite{li2023artificial,liu2023summary}, refining wearable device monitoring, and more. With the continued development of AGI, various aspects of healthcare have witnessed notable advancements. Research findings increasingly underscore AGI's capacity to usher in a transformative era for intelligent healthcare solutions. 

AGI has demonstrated its potential in enhancing patient care through the prediction of disease progression, identification of potential complications, and optimization of treatment strategies \cite{li2023artificial}. Leveraging electronic health record (EHR) data, machine learning models can forecast patient outcomes, furnishing insights critical for personalized treatment planning. Through EHR analysis, AGI can construct comprehensive patient profiles, amalgamating medical history and current health conditions. Rajkomar et al. \cite{rajkomar2018scalable} demonstrated the high accuracy of outcome prediction through machine learning using EHR data, culminating in enhanced patient care strategies. Furthermore, AGI's pivotal role lies in furnishing tailored treatment recommendations that align with each patient's distinct characteristics and requirements. Zhang et al. \cite{zhang2015paving} highlighted the effectiveness of machine learning-driven recommendation systems in providing personalized treatment plans. More recently, Venkatasubramanian et al. \cite{Venkatasubramanian2022} presented a solution for monitoring high-risk Maternal and Fetal Health (MFH) using IoT sensors, data analysis for feature extraction, and a Deep Convolutional Generative Adversarial Network (DCGAN) classifier. It continuously monitors clinical indicators like heart rate, oxygen saturation, blood pressure, and uterine tonus, and the proposed system effectively classifies MFH status into more than four possible outcomes, showing that IoT-based mobile monitoring of MFH for pregnancy care is practical. 

AGI’s potential extends to wearable health devices, enabling real-time processing and analysis of data to detect anomalies in vital signs. The capacity for on-device data processing ensures timely alerts and recommendations. Wearable devices endowed with AGI capabilities excel in detecting irregularities in patients' essential indicators, such as heart rate, blood pressure, and oxygen saturation, facilitating prompt interventions \cite{zebin2018human}. Kumar et al. \cite{9752063} developed a Smart Healthcare System (SHS) by integrating the IoT with AGI. Millions of devices and sensors capture data for continuous patient health monitoring. This data is analyzed using machine learning and deep learning algorithms to predict disease severity, and the insights are wirelessly shared with medical professionals for appropriate recommendations. 

Telemedicine and telesurgery platforms equipped with AGI offer personalized, real-time feedback and support to patients. Virtual health assistants powered by AGI can interact with patients in natural language, addressing queries, guiding them through treatment plans, and providing information and recommendations \cite{miner2016smartphone}. By integrating AGI, telemedicine platforms can also improve patient engagement and adherence to treatment plans \cite{info:doi/10.2196/jmir.7126}. Meanwhile, LLMs, a rising trend in AGI, will revolutionize how patients and clinicians access and obtain information. It is crucial for telehealth clinicians to understand LLMs and recognize their potential and limitations \cite{doi:10.1177/1357633X231169055}. With a telesurgery system powered by AGI medical robotic systems, like the da Vinci Surgical System from Intuitive Surgical (Sunnyvale, CA, USA), more complex surgeries can be performed remotely to reduce the imbalance in medical resources across geographical areas. AGI methods like GANs are also pivotal in filling knowledge voids and speeding up the incorporation of telemedicine and telesurgery into clinical practice \cite{article9}, especially for sim2real transfer learning to bridge the domain gap between simulated and real data in the development of data-driven models for medical segmentation and detection tasks that require human labeling \cite{gao2023synthetic, sahu2020endo}.

AGI-integrated smart pill dispensers manage medication for patients effectively. By learning from patient medication adherence patterns, AGI can alert patients or their caregivers in case of missed doses and predict potential health issues due to non-compliance. Johnson et al. \cite{article6} explored the application of AGI in medication management, highlighting its potential to improve medication adherence.  

With any future advancements in AGI, the healthcare sector is poised for significant transformations that will benefit both patients and healthcare providers.  




\subsection{Environmental Monitoring}

\subsubsection{Background of Environmental Monitoring with IoT} \label{sec:envi_monitor_back}
IoT technologies have been widely used for monitoring environmental conditions such as air temperature, large-scale sea surface temperature, soil moisture, air quality, etc. According to the nature of the sensor platform,  real-time environmental observation data is usually collected by various sensors including satellite-based sensors, airborne sensors, temporal in situ sensors, and long-term sensors installed at monitoring stations. 

The data collected from various environmental monitoring methods have their own advantages and disadvantages in terms of spatial resolution, spatial coverage, temporal resolution, and temporal span. 
For example, accurate and real-time soil moisture (SM) estimates are useful to characterize trends in the global and local climate systems, and for predicting the interactions between land and atmosphere \cite{guevara2019downscaling}. 
Soil Moisture Active Passive (SMAP)\footnote{\url{https://smap.jpl.nasa.gov/data/}} is an Earth satellite mission that measures and maps Earth's soil moisture by the National Aeronautics and Space Administration (NASA). SMAP data can provide global-scale soil moisture radiometer data but with a rather lower spatial resolution (36 km) and lower temporal resolution (every 2-3 days). In contrast, high spatial-resolution soil moisture data can also be collected from airborne sensors ($\sim800$ m), long-term in situ sensors (3 - 5 km), and temporal in situ sensors ($\sim5-10$ cm) such as soil moisture data measured during the joint NASA-United States Department of Agriculture (USDA) soil moisture validation campaigns for SMAP
airborne scale ($\sim800$ m), sub-pixel scale (3 - 5 km), and point scale ($\sim5-10$ cm) such as soil moisture data measured during the joint NASA-United States Department of Agriculture (USDA) soil moisture validation campaigns for SMAP \cite{gaur2016land,gaur2019nomograph}. However, these observations can only be collected by request with small spatial coverage (e.g., airborne data, temporal in situ sensor data), or have rather sparse spatial distribution (e.g.,  data from long-term in situ monitoring stations). In order to provide real-time high-resolution soil moisture observations over a large spatial scale, the best approach is to integrate observation data collected from various sensors.

\subsubsection{Foundation Models for Earth and Environmental Monitoring}

Recently, significant efforts have been made to develop FMs for climate and weather forecasting based on various environmental observation data. For example, ClimaX \cite{nguyen2023climax} is a recently developed FM for weather and climate science which are trained using heterogeneous climate, environmental, and earth observation data including 6 atmospheric variables at 7 vertical levels, 3 surface variables, and 3 constant fields. ClimaX shows promising performance on various weather global/regional forecasting, sub-seasonal to seasonal prediction, climate projection, and climate model downscaling tasks. 

In addition, IBM recently released its newest geospatial foundation model on the open-source AI platform Hugging Face \footnote{https://newsroom.ibm.com/2023-08-03-IBM-and-NASA-Open-Source-Largest-Geospatial-AI-Foundation-Model-on-Hugging-Face}. This geospatial FM was first pre-trained on NASA's Harmonized Landsat Sentinel-2 satellite data (HLS) over one year across the continental United States and was further fine-tuned on labeled data for flood and burn scar mapping tasks, two typical tasks in remote sensing domain. The results show that this geospatial foundation model leads to a 15 percent performance improvement over the state-of-the-art model by using only half-labeled data.

\subsubsection{Challenges in FM Design for Environment Monitoring}
Despite the recent success in foundation model development for earth observation and environmental monitoring, we also identify several unique challenges:
\begin{enumerate}
    \item \textbf{Integration of data with various spatial/temporal coverage:} As we discussed in Section \ref{sec:envi_monitor_back}, environmental monitoring data collected from different sensors can have different spatial coverage and temporal coverage. How to integrate them into a unified format so that they can be used for foundation model training? CLimaX \cite{nguyen2023climax} lists this as one of their major challenges when developing a foundation model based on different earth observation data. They partially solve the diverse spatial coverage challenge by leveraging the image patch idea from Vision Transformer (ViT) \cite{dosovitskiy2021vit}. They splitted the globe space into various spatial patches. For an environmental monitoring variable with a partial spatial coverage, they can only feed the patches with available data to ViT which does not necessary to form a complete grid. Similar practices can be used to solve the diverse temporal coverage challenge.

    \item \textbf{Integration of data with various spatial/temporal resolutions:} Similarly, death observation and environmental monitoring data collected from different sensors inherently have different spatial resolutions and temporal resolution. Most existing approaches \cite{nguyen2023climax,he2021spatial,saharia2022palette} simply downsample the high-resolution data or upsample the low-resolution data to make the shape of input data match each other. This practice can bring unnecessary negative impacts to the model: 1) downsampling the high-resolution data will lead to information loss and 2) upsampling low-resolution data with the bilinear spatial interpolation method will lead to data bias and errors. A resolution-agnostic architecture is preferable in this context such as some recent implicit neural representation models \cite{mai2020multiscale,chen2021liif,sitzmann2020implicit,mai2023sphere2vec}.

    \item \textbf{FM with data in various spatial data formats:} Satellite- and airborne-based environmental monitoring data are usually in the form of imagery while in-situ data are usually stored as a set of point observations. Different spatial data formats require different spatial representation learning modules \cite{mai2022review,mai2023spatialrl} so that they can be simultaneously learned by one foundation model. So far most of the multimodal foundation models focus on handling text, images, and video modalities while ignoring the importance of integrating data in different spatial formats such as points, polylines, polygons, networks, and so on. However, this is an unavoidable challenge for FM design for environmental monitoring.

    \item \textbf{FM training with limited historical data: } Unlike language foundation models which have a massive amount of data for model pre-training, the size of historical environmental monitoring data at any given time is fixed and it increases at an almost constant rate \cite{nguyen2023climax}. For example, ClimaX \cite{nguyen2023climax} utilizes atmospherical observations from 1850 to 2015 with 6-hour spatial resolutions. Such limited data size is not enough for foundation model training. To solve that, ClimaX proposes to use simulation data from various earth system models in foundation model pre-training and use real-world data on model fine-tuning. However, such kind of simulated data is only available for a limited set of environmental variables. We still need other approaches such as data augmentation to increase the size of model pre-training dataset.
\end{enumerate}


\subsection{Smart Agriculture}
Precision and smart agriculture integrates diverse technologies to enhance the productivity, efficiency, and sustainability of the farm-to-market journey. This involves capturing critical data about soil conditions, crops, and pests, which then informs comprehensive monitoring from planting to harvest. The amalgamation of data from multiple sensors, coupled with IoT devices like drones and ground robots, holds the potential to optimize resources, increase yields, and minimize costs when seamlessly interconnected. Thus, unlocking the full capabilities of precise agriculture hinges on a robust interconnectivity framework that facilitates smooth data exchange among field devices and cloud-based facilities for storage, analysis, and decision-making.

At present, local farm connections often rely on Wi-Fi or Bluetooth for short-range wireless communication \cite{feng2019study}, while remote functionalities utilize 4G cellular networks. While these solutions provide cost-effective connectivity, emerging applications in precise agriculture necessitate attributes like elevated data rates, reduced latency, and high-density communication. Consider, for instance, unmanned tractors executing precision plowing guided by GPS and computer vision \cite{rondelli2022review, kurtser2023rgb}; robots requiring real-time coordination to avert collisions and enhance cooperative planning; drones and ground robots accomplishing tasks in complex environments, relying on prompt operator feedback; and myriad sensors necessitating continuous communication for data aggregation \cite{homaei2019enhanced}. 

Farming systems are complex amalgamations of interdependent components that drive profitability, efficiency, and sustainability. Effective management of outdoor cropping systems hinges on the meticulous control of water supply via irrigation, addressing nutrient deficiencies with mineral and organic fertilizers, managing insect pressures through scouting and chemical interventions, and combating weeds using chemical and/or mechanical methods. Additionally, crucial weather conditions must be constantly monitored. Historically, these soil, plant, and environmental factors required labor-intensive, frequent visits to agricultural sites for manual data collection and sensor data retrieval. However, recent advancements in technology, including wireless data transmission, storage, and computation, have paved the way for real-time access to farm data \cite{qureshi2021sustainable}. Still, the vision of smart farming necessitates a robust network capable of accommodating multiple sensors generating substantial data volumes, demanding the high-speed data processing and transmission capabilities offered by 5G or even 6G wireless connections.

The interconnectivity solutions address key biotic and abiotic factors influencing farming systems, offering solutions across various applications. This entails integrating thermal camera systems and soil moisture sensors to detect water stress for informed irrigation management \cite{quebrajo2018linking}. Automated insect trap camera systems will be employed to monitor pest pressure and guide control strategies. Multispectral sensors will aid in identifying nutrient deficiencies and detecting weeds. All these elements can be linked to a cutting-edge 5G or 6G wireless system. Furthermore, multiple sensor platforms can incorporate sensors onto fixed mounts, as well as multi-robot ground and aerial platforms \cite{lu2021extending}.

The adoption of multi-robot collaborative Simultaneous Localization and Mapping (SLAM) \cite{lu2023bird}\cite{lu2023deep} technology holds immense promise, particularly in agricultural production and environmental monitoring. In agriculture, diverse robots can be equipped with distinct sensors to perform specific monitoring tasks, enhancing overall efficiency. For effective collaboration, real-time communication among robots (both ground and aerial) is essential, necessitating swift interconnectivity. For instance, aerial robots can capture imagery to build crop field maps and plan paths for ground robots, boosting efficiency in large fields. This entails efficient communication infrastructure to cater to multiple users, including robots and human operators. Robotic arms mounted on aerial and ground robots can be used for specialized actions such as harvesting.

In essence, IoT in agriculture is to establish robust, high-speed, and responsive interconnectivity systems, fostering multi-sensor and multi-platform communication across farms. This connected smart farming concept encompasses data from various sensors, algorithms translating data into actionable decisions, and robotic platforms implementing these decisions for profitable, sustainable, and efficient farm management. This holistic approach is underpinned by an interconnected IoT system that facilitates data transmission, processing, intelligence, and the management of robotic fleets.


Smart agriculture is a management concept implemented with advanced technology, such as big data, the cloud computing, artificial intelligence, robotics and the internet of things (IoT) for monitoring, tracking, automating, and analyzing agricultural operations \cite{Kumar2022Artifical}. The IoT is one of the important technologies in smart agriculture that can collect data and connect devices. It can provide a diverse set of tools for farmers to address several challenges in the field \cite{Sinha2022Future}. Farmers can remotely access and manage their farms from anywhere at any time using IoT technologies. The utilization of cameras and sensors first collects valuable data and then uploads it to the cloud. After data analyzing on the cloud, actuators are used to regulate farming processes automatically. Farmers can achieve these data using smart phone or PC to monitor the farm. Many IoT applications in smart agriculture have been studied in literature.
\subsubsection{Irrigation management} 
\label{sec:envi_monitor_back}
The irrigation system collects the data from soil, climate and plant using IoT technology to calculate the water requirements of crops and adjust the water flow to avoid water waste \cite{Abioye2020A}. Several IoT-based irrigation systems have been developed, such as sprinkler irrigation \cite{Barbade2021Automatic,Habib2022Design}, drip irrigation \cite{Jain2023Experimental,JainDesign} and capillary irrigation \cite{He2019Humidity}. Kumar et al. \cite{Kumar2023Evaluation} developed an IoT-based drip irrigation system comprising of capacitive soil moisture sensor, DHT11 ambient temperature and humidity sensor, DS18B20 soil temperature sensor, ESP-32 microcontroller, ESP8266 Wi-Fi module, pump, solenoid valve and solar panel. Similarly, Jenitha et al. developed a  Deep Bi-directional Long Short-Term Memory (DBLS-TM) algorithm to predict the soil moisture and rainfall status according to the air temperature, humidity, soil moisture, rainfall status, and wind speed collected by IoT sensors \cite{Jenitha2023Intelligent}. The volume of irrigation can be calculated by predicted soil moisture, rainfall status and evapotranspiration level.
\subsubsection{Fertilizers and Pesticides management} 
\label{sec:envi_monitor_back}
Fertilizers can provide all nutrient requirements of plants, crops, and soil fertility \cite{Shaikh2022Recent}. Farmers can precisely know the exact amount of nutrients required for their crops to save fertilizers and minimize harm to the ecosystem caused by excess fertilization using IoT-based fertilization system. Swaminathan et al. \cite{Swaminathan2022IoT} collected the data from optical sensors, weather measuring sensors and nitrogen phosphorus and potassium (NPK) sensors to predict fertilizer recommendation. The trained Bi-LSTM prediction model performs effectively and produces better outcomes, which is close to the expert's advice. At the same time, pesticides are also important during crop management, which can reduce the impact of weeds and pests and improve crop productivity  . To achieve precision weed spraying, Mary et al. \cite{Mary2023IOT} developed an IoT-based weeding robot equipped with an ESP32 AI camera to capture the photo in the field, a NodeMCU to detect the weed and a servo motor to control the pump of the nozzle for precision spraying. TFL Classify, a Real time image classification powered by TensorFlow Lite, was adopted to detect weeds. For pest control,  Azfar et al. \cite{Azfar2023An} set infrared (IR) light wall around the cotton plant. Once the insect obstructed the light, the position-sensitive detector can catch the light deviation. The detection coordinates were sent to the drone to respond by spraying pesticide in the detection region.
\subsubsection{Microclimate management} 
\label{sec:envi_monitor_back}
Greenhouse production is considered as an ultimate solution for increasing food demands spurred by the growing population. Greenhouse offers a year-round production environment for fresh vegetables, boasting a production rate approximately 50\% higher than open-air cultivation \cite{Ullah2022An}. Compared to other agricultural industries, the commercial greenhouse consumed largest energy \cite{Vadiee2014Energy}. Therefore, it is important to manage the microclimate precisely in the greenhouse to save energy and cost. Ullah et al. \cite{Ullah2022An} collected temperature, humidity, CO2 concentration, solar radiation and wind speed using sensors to monitor the microclimate in the greenhouse. The developed optimization scheme can be used to control seven greenhouse actuators (heater, chiller, dehumidifier, fogging system, CO2 generator and forced and natural ventilation) and achieve a tradeoff between energy consumption and plant growth rate.
\subsubsection{Plant stress detection} 
\label{sec:envi_monitor_back}
Detection of plant stress at early stage is important for improving the crop production because plant stress can lead to crop diseases and death \cite{Karthickmanoj2021Automated}. The IoT-based systems can provide an efficient and smart solution to detect plant stress\cite{Puengsungwan2018IoT,Numajiri2021iPOTs}. 
To detect the rice nitrogen stress, Zhu et al. \cite{Zhu2022Improving} developed a de-striping convolution neural network to remove strip noise in hyperspectral image and designed a nitrogen diagnosis CNN to detect nitrogen stress for rice leaves. In addition, Elsherbiny et al. \cite{Elsherbiny2022A} combined CNN and LSTM to fuse the RGB images, the weather-related factors and soil moisture collected by IoT sensors to detect the water stress. The model trained by multimodal data performed better than that by only RGB images.
\subsubsection{Livestock monitoring} 
\label{sec:envi_monitor_back}
Precision livestock farming involves automated remote detection and monitoring of individuals for animal health and welfare through the analysis of images, sounds, locations, weight and body condition \cite{Benjamin2019Precision}. It can detect issues in time and even predict potential issues based on historical data \cite{Neethirajan2021Digital}. Many measurements related to the health of animals are based on physiological responses, such as body temperature, heart rate and respiration \cite{Fuentes2022The}. To measure heart rate, breathing rate, and oxygen saturation of dairy cow, Salzer et al. \cite{Salzer2022A} designed a nose ring sensor by integration of thermal and photoplethysmography sensors. Besides, some studies mounted accelerometer, IMU and GNSS devices to monitor the behavior of animals\cite{Arablouei2023Animal,Shahbazi2023Deep,Fujinami2023Evaluating}. Arablouei et al. \cite{Arablouei2023Multimodal} fused the accelerometry and GNSS data to classify animal behavior. The performance of posterior probability fusion is preferable to that based on feature concatenation. However, many sensors are contact or invasive, which may cause potential pressure on animals. Therefore, many studies are focused on contactless methods, such as computer vision \cite{Wang2022The,Fuentes2022Non,Jorquera2019Modelling,Bloch2020Automatic}. Guo et al. \cite{Guo2022Development} employed object detection methods to obtain the eye temperature from thermal image of the eye socket in 3 seconds.

AGI can be a powerful and efficient solution for IoT applications in agriculture. AGI-driven IoT system can adaptively regulate irrigation by analyzing multimodal data, such as the soil moisture, environment data and crop needs to optimize water usage and reduce water waste \cite{Lu2023AG}. AGI can evaluate the soil and crop conditions as well as weed and pest threats through IoT-sensors to recommend precise dosages of fertilizers and pesticides, which can minimize overuse, maximize crop health, and reduce environmental impact. AGI-powered microclimate management systems can create optimal conditions for crops in greenhouse by controlling factors like temperature, humidity, light, and gas to improve the plant growth and productivity. In plant pressure detection, AGI can process various datasets from IoT sensors to detect early signals of plant stress caused by diseases and pests. Growers can enable proactive interventions to prevent crop losses and financial costs. In precision livestock farming, AGI can analyze data from wearable IoT devices and contactless sensors to monitor the health, behavior, and productivity of individual animals. This enables early disease detection and optimized animal welfare. However, there are also some challenges when using AGI in IoT for agriculture. The agricultural environment is complex, a robust and reliable AGI-driven system is necessary. Furthermore, some agricultural applications, such as spraying and irrigation, require real time data processing, which could be challenging for complex AGI models. Besides, the diverse range of IoT devices used in agriculture may cause compatibility issues. Seamless integration of the devices from different manufacturers and standards can be a challenge. Finally, protecting the vast data from cyberattacks is essential to maintaining farm security and animal welfare.

\subsection{Smart Transportation}

\subsubsection{Intelligent Motor Drives}

The majority of recently proposed detection methods can be classified as either physics-based methods or data-driven methods.
Physics-based methods commonly detect cyber-attacks by analyzing pre-defined system performance metrics or residuals between predicted system variables and corresponding true measurements~\cite{giraldo2018survey}. However, most physics-based methods rely on accurate physical models of the target systems, which are unavailable for most cyber-attack scenarios. In real-world applications, cyber-attacks are highly unpredictable, and their analytical impact models heavily depend on specific attack policies. These factors render the performance of most physics-based methods unreliable.
Recent research has begun to harness the power of data-driven methods to develop model-free detection methods in power electronics systems, reducing dependency on physical models ~\cite{habibi2020detection} adopted a specific type of RNN, namely a nonlinear auto-regressive exogenous model, to detect false data injection attacks in microgrids.~\cite{dehghani2021cyber} proposed an attack detection method by combining deep neural networks and wavelet singular value decomposition.~\cite{khan2021intelligent} employed multi-class support vector machines to detect and localize false-data-injection and denial-of-service attacks in inverter-based systems.~\cite{li2021detection} proposed a detection and diagnosis method targeting data integrity attacks in solar farms using a multilayer long short-term memory network.~\cite{li2020data} examined the effectiveness of various standard data-driven methods with micro-PMU data in detecting cyber-attacks in PV farms.~\cite{yang2022data,yang2022fast} developed anomaly detection methods for electric vehicle traction motor drives using a combination of support vector machines, random forests, k-nearest-neighborhood, and logistic regression.~\cite{yang2022datadriven} employed supervised classification methods to differentiate cyber-attacks and physical faults in manufacturing motor drives.
Despite the advantages of recently developed data-driven approaches, a significant challenge in deep learning-based cyber-attack detection in power electronics systems is the requirement for large-volume training datasets. The model can only learn features incorporated in the training data, and the algorithm may fail when testing data contains different features~\cite{8909793,8058000}. To address this issue, a large-scale training dataset is necessary to include similar data to the testing data. However, the computational cost due to the large volume of training data hinders deep learning model performance. Transfer learning techniques have been proposed to enable machine learning models to leverage knowledge from one domain to another~\cite{9328227}, thus reducing the amount of required training data~\cite{9773401}. Deep transfer learning methods have been utilized in cyber-attack detection and fault diagnosis in intelligent machine systems. Some methods~\cite{9806715,8909793,8758199} employ deep adversarial models to achieve transfer learning by minimizing predicted domain labels, while others aim to minimize the discrepancy between learned features from the source and target domains~\cite{8058000}.


\subsubsection{Connected and Autonomous Vehicles}
Emerging smart transportation technologies such as connected and automated vehicles (CAVs) provide new opportunities to improve safety, energy efficiency, sustainability, and mobility of the automotive and transportation sector \cite{Shao_VTM, Shladover_review, Vahidi_review}. Connected vehicles (CVs) are essentially \textit{edge IoT devices} that can connect to the internet or other \textit{edge devices} such as surrounding CVs, smart infrastructures (e.g., traffic signal lights), and other connected road users (e.g., pedestrians, cyclists). CVs are equipped with onboard communication devices that are enabled by communication technologies \cite{cv2x_dsrc_1, cv2x_dsrc_2} such as 5G, C-V2X, or dedicated short-range communication (DSRC). Thus, real-time communication can be established to obtain information on other vehicles' position and speed, signal phase and timing (SPaT), routing and road curvatures, etc. In addition to connectivity, a CV equipped with vehicle automation technologies becomes a CAV whose real-time vehicle states and motions can be precisely controlled. SAE has defined different levels of automation \cite{SAE_J3016}, from advanced driver assistance systems (ADAS) such as adaptive cruise control (ACC), lane keeping, to fully automated vehicles. CAV technologies not only leverage real-time communication to significantly extend the line-of-sight of a human driver or traditional onboard sensors to receive traffic information hundreds of meters away \cite{9406371}, but also bring in new control means through vehicle automation. Therefore, future traffic conditions can be predicted more precisely to provide opportunities for proactive vehicles or traffic control strategies \cite{Shao_VTM}. 

Many CAV technologies have been studied in the literature. It is necessary to give an introduction to the state-of-the-art CAV applications before discussing how AGI can potentially address challenges in the current literature. CAV technologies can be categorized into the following levels based on control means:

\paragraph{Powertrain} A vehicle's powertrain control includes energy management strategy (EMS) and transmission gear shifting control. EMS determines the battery-engine power-split of hybrid electric vehicles (HEVs), power-split of hydraulic hybrid vehicles (HHVs), and multi-motor power-split of electric vehicles (EVs). Transmission gear shifting control changes the torque and speed ratios between the engine and the driveline \cite{Shao_gear}. Powertrain control strategies in production vehicles are often tuned conservatively for the most demanding conditions to balance between efficiency and drivability. They are mostly rule-based and cannot adapt to varying driving conditions \cite{8505416, QIN20181627}. Many smart powertrain control approaches have been studied in literature to leverage better predicted future conditions from CVs. Therefore, powertrain efficiency can be improved by proactively controlling gear shifting of conventional vehicles or power-split of HEVs \cite{Shao_HEV, Azrin_VT, Azrin_TRC, Rizzoni_book, Huei_HEV},  HHVs \cite{FengWang_offroad, FengWang_2, 6507615}, and EVs \cite{Jin_EMS, trovao2016energy, Shao_EV}. For example, EMS can use more battery power when it anticipates a HEV will decelerate during an intersection as the battery power will be replenished from regenerative braking \cite{Shao_DSC_magazine}. 

\paragraph{Individual vehicle} On top of powertrain control, a vehicle can be better controlled by leveraging CV information. First, vehicle speed can be optimized based on predicted future traffic conditions. For example, a CAV can receive future SPaT to arrive at an intersection during green light and avoid waste due to braking. This is referred to as \textit{Eco-approach} applications \cite{Shao_eco-approach, 8481424, 10.1115/DSCC2018-9059}. Second, vehicle routing can be planned to reduce travel time based on real-time information of the traffic flow speed \cite{doi:10.3141/2619-01, 8960401}. 

\paragraph{Multi-vehicle cooperative driving automation (CDA)} An individual vehicle-based vehicle control strategy can be selfish and compromise the performance of other vehicles. CDA strategies can be developed to optimize multiple vehicles to achieve global optimality for both the overall traffic and each vehicle (agent). Applications such as cooperative adaptive cruise control \cite{8569947, doi:10.3141/2489-17, 4019451, 7963723}, cooperative merging \cite{7534837, NTOUSAKIS2016464, doi:10.1061/JTEPBS.0000243, Shao_merging}, speed harmonization \cite{GHIASI2019210, 8464283} have been proposed in the literature. In addition, when considering multiple CAVs, cooperative perception \cite {9454591} becomes possible where CAVs collaboratively perceive the environment to extend perception beyond local sensing capability, improve safety, and reduce the computational power needed.

\paragraph{Traffic infrastructure} In addition to vehicle-centered control strategies, traffic infrastructures can be further optimized as CVs provide real-time traffic information regarding arrival time, demand, routing, etc. The literature has studied various applications, including smart signal control strategies \cite{GUO2019313, 9708714, doi:10.1177/0361198118790860}, variable speed limit \cite{HAN2017113, doi:10.1061/JTEPBS.0000379}, ramp metering \cite{9198911, doi:10.1061/9780784481547.004}, etc. Also, because of IoT, CVs or connected infrastructures can be used as remote sensing devices to provide real-time traffic data. Proactive traffic management strategies \cite{abdel2019integrated, hadi2019connected} can be designed to improve overall traffic performance, such as lane management, incident and emergency response, and provision of information and guidance. 

It is anticipated that combining the above categories can introduce further benefits. For example, both individual and multi-vehicle control strategies can be integrated with powertrain level control to improve performance for CAVs \cite{Shao_HEV, Shao_eco-approach}. Co-optimized vehicles and traffic infrastructure control, such as integrated vehicle speed and signal light control, can avoid conflicts between the two systems and maximize energy efficiency and mobility \cite{YANG2021102918, 8408521}. Recently, the electrification of vehicles and transportation has attracted more attention, and several new opportunities have been proposed in the literature. EVs have different powertrain architectures and can be controlled to improve powertrain efficiency. Electrified vehicles generally have a faster response time to effectively control and actuate the vehicles in highly dynamic driving conditions. EVs also have more electrical power onboard to support increased computational capabilities for complex driving tasks. 

AGI can be a promising solution for the above CAV and smart transportation applications in the literature. All the above applications require an effective prediction of future traffic conditions, which relies on the modeling and understanding of the overall system. The system can include several vehicle types with clearly different driving behaviors: human-driven vehicles, CVs with human drivers, individual CAVs, and cooperative CAVs. Traffic and driving behaviors are complex to model and highly nonlinear and stochastic in nature. It is challenging to propose a generic model as traffic dynamics depend on the specific road geometries, whether it is signalized, curvature, speed limit, number of lines, traffic movements and approaches, etc. Even for traffic on the same road, the conditions can change drastically according to the time of day, events, or incidents. A small change in the road geometry can greatly impact traffic patterns and reduce the accuracy of an already calibrated and trained model using either traditional methods or AI-based approaches. Novel AGI techniques can significantly improve the accuracy of traffic modeling and prediction and obtain generic models that can adapt to the conditions and variants mentioned above. Breakthroughs in traffic modeling using AGI can also benefit the evaluation of CAVs. The fidelity of simulation will be greatly improved to reduce the amount of real-world testing that is expensive, time-consuming, and causes safety concerns \cite{shaorealsimosti}. In addition, traffic models are usually highly nonlinear and complicated, which makes it challenging to solve the associated control or optimization problem in real-time. AGI can help identify an efficient model structure and be used to effectively search the optimal control strategies for real-time implementation. This is especially promising for autonomous driving tasks \cite{Hu_2023_CVPR}, which are traditionally divided into tasks, including localization, perception, prediction, planning, and motion control. A unified end-to-end AGI approach can greatly reduce the accumulative errors from each module and improve coordination among different tasks. Such an AGI-based autonomous driving model has already attracted much attention in academia and industry \cite{DBLP:journals/corr/BojarskiTDFFGJM16, Prakash_2021_CVPR, tesla_end2end}.

\subsection{Smart Manufacturing}

The Industry 4.0 led by organizations such as GE and IBM has allowed better production process description, communication, and computation, and this trend is being accelerated by the mass adoption of IoT devices, 5G communication technology, the advancement of AI and AGI technologies~\cite{lasi2014industry,zheng2021applications,bu2021iiot,badini2023assessing}. Accordingly, the Strategy for American Leadership in Advanced Manufacturing (AM) states that worldwide competition in manufacturing has been dominated in recent decades by the maturation, commoditization, and widespread application of computation in production equipment and logistics. 


With the IoT devices
,  various modeling, monitoring and control approaches have been proposed for manufacturing quality, performance and reliability improvements~\cite{montgomery2009statistical,shi2006stream,arinez2020artificial,sun2020cyber}. For instance, the product geometric accuracy was investigated in~\cite{huang2015optimal} and different types of process properties, such as porosity, roughness, were investigated in 3D printing with various process sensing information from IoT sensors~\cite{sun2017quality,sun2017functional,li2020manufacturing}. The model sparsity and interpretability was considered during the sensor data quality modeling in semiconductor manufacturing~\cite{sun2016logistic,sun2020supervised}. To address high dimensional sensor and video data, tensor techniques and deep learning models are  investigated in smart manufacturing~\cite{huang2020geometric,huang2020unsupervised,segura2020nearest,segura2023droplet}. In terms of IoT data monitoring and anomaly detection, methods such as binary segmentation, dynamic programming, Bayesian methods, nonparametric methods, and model-driven methods were proposed
\cite{kamat2020anomaly,lindemann2019anomaly}. For instance, Bayesian online change detection was used for the droplet video monitoring during inkjet printing~\cite{segura2021online}. Workers play a significant role in manufacturing, warehouse and other systems. The consideration of human safety and human performance and their impact to the general system performance in engineering systems was  investigated, see for instance~\cite{hajifar2021forecasting,kheiri2023human,vahedi2023relationship,lamooki2022data,hajifar2021investigation}.

The digital tools in IoT and Industry 4.0 enabled the remote and connected  tracking and control, at any place and platform, by connecting the once isolated Operational Technology (OT) networks to the rest of the enterprise through the Information Technology (IT) network~\cite{prinsloo2019review, ani2017review}. While this allows for a better manufacturing modeling, monitoring and control, it also poses profound questions in areas such as intellectual property (IP) protection, data security and privacy. These issues have led to product recalls, public backlash, lost revenue for companies, and will potentially threaten public safety and security~\cite{Deloitte2019,ISACA2021}. Specific techniques were proposed to address the above-mentioned challenge. Differential privacy and privacy-preserving modeling are proposed to address the privacy and security issue~\cite{hassan2019differential,huong2021detecting,liu2020deep,wang2021toward}. For instance, G{\'o}mez et al. integrated traditional differential privacy with k-anonymity, so that the preserved data can be aggregated and transmitted without risking the privacy 
\cite{gomez2016using}. 
Federated Learning (FL) was used so that machines collaboratively train a model without sharing raw data~\cite{liu2020deep,wang2021toward}. For instance, Liu et al. proposed a FL monitoring framework to enable decentralized edge devices to collaboratively train an anomaly detection model~\cite{liu2020deep}. Wang et al. applied FL to build a universal anomaly detection model with each local model trained by the deep reinforcement learning~\cite{wang2021toward}. However, it is reported FL suffers from data leakage~\cite{jin2021cafe,ren2022grnn}. 

Though the machine learning and AI approaches have shown useful in addressing a certain type of problem, it can be hard for them to work in general problems with varying and heterogeneous environments, limited annotated data, multimodal information source, etc. AGI poses opportunities to address these challenges, and has the potential to provide more comprehensive and robust solutions to further enhance the capabilities of existing AI based methods for manufacturing modeling, monitoring and control. \cite{kumpulainen2022artificial} surveyed the AGI and Industry 4.0 fields, and found that though AGI studies has huge potential, the gap between the AGI studies and the industry needs is high.

We point out the potential of AGI in manufacturing here. 
\begin{enumerate}
    \item \textbf{Manufacturing system heterogeneity and model generalization:} The manufacturing processes and systems can be complex with heterogeneous machines and operations, and most AI methods may not be able to cope with the heterogeneity. There is a need to develop models and methods that is generalizable to various scenarios. AGI has the potential to adapt to new scenarios, by  transfer learning from one domain to another \cite{toscano2021deformation,kontar2020minimizing} or incorporating physics driven domain knowledge \cite{niaki2021physics,hajiha2022physics}. For instance, \cite{toscano2021deformation} investigated the feasibility of transferring the process knowledge at various manufacturing scales. \cite{hajiha2022physics} proposed a framework that consists of a data layer and a physics layer, to capture the statistically-correlated temporal dynamics and  imposes regularizations through system working principles and governing physics, respectively. More endeavors are needed to deal with the manufacturing heterogeneity and achieve generalizable models.

    \item \textbf{Manufacturing data augmentation and labeling, and learning with limited data:} The manufacturing data collection and labeling can be expensive. AGI has the potential to deal with the challenge, by performing data augmentation~\cite{toscano2023teeth,chung2023anomaly}, adaptive/interactive data labeling and annotation~\cite{li2023multiclass,li2023reinforced}, and learning with limited data~\cite{zhou2020siamese,russell2023maximizing}. GAN has been used to generate useful samples for teeth aligner printing and point cloud characterization~\cite{toscano2023teeth}, and data augmentation for supervised anomaly detection in additive manufacturing~\cite{chung2023anomaly}. Generative models have been used in CAD software for engineering designs \cite{gerhard2022generative}. Reinforcement learning and active learning can adaptively identify the important samples to annotate and achieve process characterization with limited annotation efforts~\cite{li2023multiclass,li2023reinforced}. 
    \cite{zhou2020siamese} used a Siamese CNN based few-shot learning network to measure distances of input samples based on their optimized feature representations, which is helpful for achieving good anomaly detection performance with limited samples.

    \item \textbf{Multimodal manufacturing data analysis:} Manufacturing system conditions can be measured and reflected with various data types and sources, such as IoT sensors, images, domain knowledge, and texts. How to make the best use of multimodal information and support decision-making is attracting much attention. While data fusion and multimodal analysis have been studied for decades \cite{jin2015ensemble,li2018integration,wang2022energy}, the solutions are generally tailored for an application scenario and can be hard to generalize. AGI has the potential to overcome this challenge. Recently, contrastive learning was studied to acquire effective data representations from multiple modalities for downstream tasks ~\cite{kipf2019contrastive,ai2022domain}. For instance, Ai et al. integrated knowledge distillation to transfer the information from handcrafted features to deep learning and supervised contrastive learning to enhance feature discrimination~\cite{ai2022domain}.

    \item \textbf{Manufacturing diagnosis and troubleshooting:} Diagnosis and troubleshooting is an important step in the manufacturing value chain, but was mainly done manually by experts and workers in the past. With the advent of tools such as LLMs, AGI has been explored for the manufacturing process and system diagnosis and troubleshooting~\cite{power2019generalized,rathore2023future,badini2023assessing}. This has the potential to bridge human knowledge and AI methods to train reliable decision-making tools. Power et al. studied artifacts such as cars and circuit designs using Non-Axiomatic Reasoning System (NARS), which demonstrated certain features of the generalized diagnostics. NARS can diagnose the abnormal states of unknown artifacts without having prior knowledge on them~\cite{power2019generalized}. The collection of manufacturing know-hows and building a reliable AGI for diagnosis and troubleshooting is promising but needs tremendous effort.

    \item \textbf{Manufacturing assistance and workforce training:} Manufacturing workforce training in the Industry 4.0 and IoT era is of super importance to national security, and there is a huge lack in the manufacturing talent pool \cite{plumanns2019organizational,li2021data}. AGI has the potential to revolutionize the way workers learn and interact with machines. For instance, AGI can facilitate interactive question answering and learning and smooth the learning curve. AGI can also help simplify the operation of machines by developing smart manufacturing assistants. These assistants could optimize the recommendations based on real-time data from the machines and adapt to individual workers’ preferences. 
\end{enumerate}


\subsection{Smart Education}
The potential of integrating AGI and IoT into education has prompted educators to reconsider traditional teaching methods and embrace innovative approaches~\cite{woolf2015ai,eysenbach2023role}. When applied to education, AI and IoT can address various challenges and enhance the learning experience. Recent developments in AGI technologies, such as OpenAI's GPT-4, have increased awareness about using digital resources in K-12 and higher education to equip students for 21st-century problem-solving\cite{baidoo2023education}. AGI tools like ChatGPT and GPT-4 enhance personalized and interactive learning by providing formative assessment prompts, feedback, and relevant references. They assist teachers with lesson planning, including content knowledge and assessment strategies, offering a wider range of content to enrich teachers and help students gain diverse perspectives. AGI frees teachers from textbook limitations and expands their content knowledge. For instance, using the 5Es model, ChatGPT designed a learner-centered teaching unit on renewable and nonrenewable energy sources, generating a rubric for student self-evaluation, exam questions, and a scoring key for teacher-led evaluation~\cite{cooper2023examining}. Key concepts arising from discussions on AI and education include improving teachers' content knowledge, facilitating individualized and adaptive learning, differentiated instruction and assessment, and enhancing educational outcomes for students. As AGI technologies progress, further research will explore their practical applications in the classroom. While the integration of AGI to bridge the gap between IoT and AGI within education remains an ongoing endeavor, the following section explores the insights from diverse scholarly works to elucidate the transformative potential of AGI and IoT within education.

\textbf{Educational Goals}. 
The attainment of educational goals within any learning environment is a critical endeavor, aimed at fulfilling designated learning objectives and desired student outcomes. Despite resource constraints, shortages of skilled educators, and inadequate attention to diverse learning needs, strides have been made with the integration of IoT technology. This integration has the potential to reshape the educational landscape by fostering interactive and personalized learning experiences. However, while IoT's integration in education offers promising opportunities, several challenges must be addressed. Data privacy and security concerns loom large, as the abundance of sensitive information collected by IoT devices exposes vulnerabilities that can be exploited by cyber threats \cite{canbaz2021iot,el2023effectiveness}. Moreover, the digital divide persists as a challenge, potentially exacerbating educational inequalities. Unequal access to IoT infrastructure and resources could lead to a disparity in learning experiences \cite{andiyan2021disruption}, hindering equitable goal achievement. Striking a balance between technological advancement and accessibility remains a critical hurdle. AGI emerges as a promising solution in surmounting the obstacles posed by IoT in educational goal attainment. Leveraging AGI's capabilities, one can develop sophisticated predictive models that identify students at risk and customize interventions based on a holistic range of factors. AGI can bolster data security through adaptive threat detection mechanisms, fortifying the protection of sensitive information. Furthermore, AGI-powered educational tools can bridge the digital divide by providing personalized learning experiences to improve the outcomes \cite{pedro2019artificial,yang2020practical,zhai2023chatgpt}, even in resource-constrained environments. Through adaptive content delivery and intelligent tutoring systems, AGI has the potential to foster inclusivity and equitable access to quality education. As AGI continues to evolve, its integration holds the promise of not only addressing IoT-related challenges but also propelling educational goals within reach, enabling learners to thrive in an interconnected world.

\textbf{Pedagogy}. 
As educators delve into the IoT realm, they encounter challenges that could hamper the realization of its full potential in education. Integrating IoT into pedagogy requires renovating instructional methods. Furthermore, not all educators are equipped with the necessary skills or training to incorporate these new tools effectively into their teaching. This has led to a growing gap between the potential benefits of IoT and its actual educational application.
AGI presents promising solutions to these challenges. By leveraging AGI's cognitive capabilities, IoT devices can be equipped with advanced data analysis and contextual interpretation abilities. This enables personalized and adaptive learning experiences for students by processing vast amounts of data generated by IoT devices, extracting meaningful insights, and tailoring instruction that addresses the individual needs of students. For example, a study exploring AI-based strategies in teaching art courses found that AI technology had significant reference significance in improving teaching effectiveness~\cite{yang2020practical}. Multi-modal AGI tools from OpenAI are able to handle various types of IoT data in order to make better assessment to students, allowing educators to accurately evaluate progress and cater to specific needs and learning styles~\cite{zhai22023chatgpt}. AGI's ability to translate text into different languages and convert it to different languages, even including sign language, improves accessibility for students with disability and non-native speakers, visually impaired students, and those with reading difficulties via IoT devices, making learning more inclusive~\cite{atlas2023chatgpt}. Additionally, the variety of resources available through OpenAI can greatly benefit teachers in learning how to utilize new IoT tools in differentiating instruction and developing targeted lesson plans and teaching strategies to support each student effectively.

\textbf{Curriculum Design}. 
Integrating IoT techniques into curriculum design represents a significant advancement in education, enabling students to understand and harness the potential of interconnected smart devices. However, this integration poses challenges across four critical dimensions: discipline-specific, learner-centered, career-oriented, and society-oriented ~\cite{chiu2020sustainable}. AGI's capacity to generate customized learning materials empowers educators to develop IoT-centric modules that align with the discipline's specific requirements. These modules are readily to include discipline-specific individualized case studies, examples, and practical applications of IoT concepts, making the integration more engaging and relevant for students, which personalizes learning experiences and demonstrates comprehensive assessments \cite{hu2020design}. Personalized learning is made possible through IoT-driven data analysis. Learning management systems can collect data on students' progress, preferences, and learning styles, enabling educators to tailor content and pacing to individual needs. AGI-powered adaptive systems can dynamically adjust learning materials, assessments, and activities based on real-time IoT data ~\cite{zhai2023chatgpt}. IoT integration can prepare students for the evolving job market by simulating workplace scenarios. For instance, in business education, students can engage with IoT-driven simulations of supply chain management, decision-making processes, and customer interactions. This hands-on experience enhances their problem-solving skills and industry readiness. IoT-enabled projects can emphasize societal challenges and encourage students to develop solutions. In environmental studies, students can design IoT sensors to monitor pollution levels, promoting awareness and proactive environmental stewardship. In social sciences, IoT data collection and analysis can provide insights into urban planning, public health, and community engagement.

\section{Challenges and Optimizations}

\subsection{Limited Computing Resources and Real-time Response}
After Deep Neural Network (DNN) models are trained with large volumes of data, they can be applied to a broad spectrum of devices, including but not limited to sensor nodes, wireless access points, smartphones, wearable technologies, video streaming devices, augmented reality systems, robotics, unmanned vehicles, and smart health devices~\cite{philipp2011sensor,lane2015early,boticki2010quiet,rodgers2014recent,bhattacharya2016smart}. 
Recent breakthroughs in transistor density have led to a significant increase in the computational power of these devices. This advance enables applications, previously limited to high-performance CPU/GPU environments, to be effectively executed on these devices. As a result, IoT devices have emerged as the preferred platform for the applications discussed in this paper, given their capabilities, enhanced privacy protections, and low power consumption.

Given the nature of the applications, achieving real-time performance (typically 30 processes per second or 33 milliseconds per processing) is a principal criterion. However, the progressively expanding size of DNNs poses a critical challenge in delivering real-time inference performance, particularly for LLMs ~\cite{touvron2023llama}. These models utilize over a thousand computational layers and more than a billion parameters~\cite{touvron2023llama, niu2021dnnfusion} in order to achieve superior accuracy. However, this poses significant challenges for deploying these models on IoT devices, especially when it comes to real-time performance. IoT devices have limited computation resources such as memory bandwidth, throughput, and power budgets. As a result, there is a complex trade-off between the increasing complexity of DNNs, which is necessary for improving accuracy, and the deployment of these DNNs on resource-constrained mobile devices to enable wider application.

From the hardware perspective, there is a growing trend in designing specialized system-on-chips (SoCs) for IoT devices, which provide a programming interface suitable for general-purpose processing ~\cite{edgetpu,mcxmcu,ethosu65}. Smartwatches, in particular, have made significant investments in advanced Apple Silicon chips ~\cite{applewatchs8}, which can also accelerate highly parallel workloads. However, there has been limited exploration of using these specialized chips for other types of workloads. In the next section, we will analyze existing efforts and potential optimizations to address the challenges mentioned earlier.

\subsubsection{Algorithm-level Optimizations}
DNN model compression techniques, such as pruning\cite{han2015learning,guo2016dynamic,dai2017nest,mao2017exploring,wen2016learning,he2017channel}, have been proposed to reduce storage and computation while accelerating inference. However, there is a trade-off with a slight accuracy loss. Pruning can be performed during DNN training and is considered an algorithmic optimization that reduces redundancy in weight numbers. There are three types of weight pruning schemes: non-structured (irregular), coarse-grained structured (regular), and fine-grained structured (regular). Non-structured pruning~\cite{han2015learning,guo2016dynamic,dai2017nest} allows arbitrary weights to be pruned, resulting in a higher pruning rate but potential performance degradation in GPU and CPU implementations due to sparse matrix storage format. On the other hand, structured pruning\cite{mao2017exploring,wen2016learning,he2017channel} leads to regular smaller weight matrices in GPU/CPU implementations, resulting in more significant acceleration. Fine-grained structure pruning~\cite{zhou2021learning, patdnn, ma2019pconv, cai2021yolobile} prunes fine-grained patterns within coarse-grained structures. The data access pattern and computational pattern are more regular, offering high accuracy, hardware-friendliness, and a high pruning rate.

Weight quantization \cite{leng2017extremely,park2017weighted,zhou2017incremental,lin2016fixed,wu2016quantized,rastegari2016xnor,hubara2016binarized,courbariaux2015binaryconnect} is another model compression technique that reduces redundancy in the bit representation of weights. It involves mapping floating-point weights to a set number of quantized levels, determined by the chosen bit representation (e.g., $n_k$-bit). Each quantized weight is equal to the weight quantization scaling factor $\alpha_k$ multiplied by the value represented by an $n_k$-bit digit.
Compared to pruning, weight quantization is considered hardware-friendly and offers proportional computation and storage requirements reductions. Consequently, it has become a prominent technique for compressing DNNs, often seen as an essential step in FPGA and ASIC designs for DNN inference engines. Furthermore, support for weight quantization has expanded to include GPUs and mobile devices such as NVIDIA GPUs with PyTorch \cite{paszke2017pytorch} and TFLite \cite{TensorFlow-Lite} support for mobile devices.

\subsubsection{System-level Optimizations}
Recently, substantial efforts have been directed towards accelerating DNN inference frameworks for edge and IoT devices. These efforts include the development of DeepX~\cite{lane2016deepx}, TFLite~\cite{TensorFlow-Lite}, TVM~\cite{chen2018tvm}, Alibaba's Mobile Neural Network (MNN)~\cite{Ali-MNN}, DeepCache~\cite{xu2018deepcache}, DeepMon~\cite{huynh2017deepmon}, DeepSense~\cite{yao2017deepsense}, and MCDNN~\cite{han2016mcdnn}. However, most of these frameworks failed to fully exploit model compression techniques, and none succeeded in real-time execution of LLMs on edge devices. Although previous attempts have been made to perform inference using compressed DNNs on edge platforms (e.g., ~\cite{vooturi2017efficient}, the interaction between model compression and hardware acceleration has not been meticulously explored in previous studies. 
There are still some open questions to further reduce the storage and computation cost for emerging LLMs.



\subsection{Large-scale IoT Communication}

As IoT devices become prevalent and deeply integrated into our daily life, it becomes crucial to establish both intra- and inter-connectivity among them. Such connectivity is vital for data transfer, sharing, and analysis. However, as the number of IoT devices grows exponentially, a grand challenge a significant challenge arises in facilitating smooth communication between them. 
This challenge is particularly pronounced in wireless communication, where interference becomes a major concern when sharing limited spectrum resources \cite{9693972}. The scarcity, limitations, and cost associated with spectrum availability make this a complex issue that cannot be easily resolved by merely allocating additional spectrum bands. 
Consequently, over the past few decades, substantial research efforts have been dedicated to enhancing communication design, such as improving spectrum and energy efficiency \cite{aslam2018energy}, enabling large-scale and high-density connectivity \cite{badi2021reapiot, xie2023edge}, supporting diverse communication needs (low latency, high data rate, high reliability), and more. 

In a classic communication system with one transmitter (TX) and one receiver (RX), TX will translate application data into binary form. These binary data then undergo a sequence of operations: channel coding, modulation, transmission, demodulation, and de-channel coding (commonly known as the classic 5-step process) \cite{tse2005fundamentals}. Eventually, they are translated back from binary to their original application data format. At each step, robust solutions have been established, primarily rooted in the framework of probability and statistics. These solutions form the bedrock of the dynamic wireless research landscape and the products we have today. 
While significant progress has been made in this area, leading to several generations of wireless technology (1G to the latest 5G), the intricate challenges arising from factors such as high traffic, heterogeneity, constant connectivity, and dynamic conditions have rendered conventional optimization-based approaches unfeasible. 
Notably, optimization at each transmission step introduces high computational cost, not to mention they are disjoint in terms of end-to-end optimization. For IoT devices, these challenges are more severe, since they typically have limited computation and power capacity. 

In recent years, ML-based approaches have emerged as dominant forces within wireless research, particularly at a lower layer of Open Systems Interconnection (OSI) model, such as physical layer (PHY). While the transmission of electromagnetic signals from transmitters (TX) to receivers (RX) involves intricate physical interactions with the surrounding environment (e.g., reflection, diffraction, and diffusion), ML tools exhibit remarkable prowess in effectively managing these random behaviors. Furthermore, ML-driven approaches possess the unique ability to bypass the traditional 5-step process and engage in end-to-end optimization. One factor that may hinder the traditional ML-based wireless system is the \emph{generalizability}.

Given the recent introduction of more robust AGI models, it's not beyond reason to anticipate that AGI could even devise efficient IoT wireless communication systems that have hitherto remained unexplored. 
For example, a model that may work well in an indoor setting but cannot work well outdoors; or a model can optimize laptop communication but cannot handle IoT devices. The contributing factors are not only coming from the limitation of traditional models (CNN, RNN, LSTM), but also insufficient datasets. Fortunately, in the age of massive IoT, we foresee the formal can be solved with a more powerful AGI model, while the latter can be addressed by a huge volume of data from pervasive IoT devices. 

In the following section, we still follow OSI layer model and provide a brief literature review, each with our observation and remark. Different from existing literature reviews and insights~\cite{hussain2020machine, aboubakar2022review}, our emphasis is the impact that new AGI models can bring to large-scale IoT wireless communication.

\subsubsection{AGI for Physical Layer}
PHY layer mainly handles raw signals, such as analog-to-digital (ADC) and digital-to-analog (DAC) conversion, radio frequency (RF), signal estimation and detection, etc. As mentioned above, PHY layer is the most active area that has been applied with ML. There have been extensive reviews on ML-based PHY techniques, such as those summarized in~\cite{wang2017deep}. However, these works mainly focused on a specific task. For example,~\cite{yuan2019machine, aldossari2019machine} discussed approaches for channel estimation, and~\cite{nguyen2017joint, zhang2020machine} introduced ML-based coding/decoding. These works may work well under their proposed scenarios but can hardly extend to a more general setting, for example, the IoT scenario. What makes IoT PHY different from other communication applications is that IoT communication reveals scalability and heterogeneity. A plausible solution is AGI-based one. The idea is to collect enough IoT wireless data, which can be relatively easy to obtain due to their ubiquity, and then use foundation models to train. The objective here is to accurately predict PHY parameters, such as transmit power, bandwidth allocation, coding scheme, etc. 

\subsubsection{AGI for MAC Layer}
MAC layer handles user coordination, such that when multiple users try to access the shared medium, it can efficiently coordinate resources, such as spectrum bandwidth, power, time, etc. MAC is of critical importance, especially for wireless communication, since wireless signals can cause interference (hence unsuccessful transmission) if not coordinated properly. In the past, scheduling is usually formulated as an optimization problem, usually non-convex~\cite{9625544, asadi2013survey}. Solving the problem is challenging and in most scenarios, the optimal solution is not possible. In recent years, ML-based methods, such as deep reinforcement learning (DRL)~\cite{wei2018drl}, prove to be very effective, especially when scheduling becomes complicated when IoT users scale. 
As a more generic approach, the powerful AGI tools can remedy this problem. Specifically, we envision that AGI can effectively tackle the non-linear binary optimization problem involved with user scheduling, and it should scale well as indicated by existing ML tools \cite{cui2019spatial}. 

\subsubsection{AGI for IP and Network Layer} 
IP and network layer process routing and network-specific traffics, respectively. Recently, it has become a research hotspot since it mainly tackles network anomaly activities \cite{chandola2009anomaly}. In \cite{ahmed2016survey}, anomaly detection has emerged as a primary challenge in IoT networks, and several techniques were proposed, with a focus on traditional fingerprint-based detection. Usually, packet IP address, length, protocol type, etc, can be applied to identify if the incoming packet is normal or abnormal. However, such rule-based methods cannot be generalized as the communication becomes more complicated and diverse. The past decade has witnessed some novel anomaly detection architects, especially those in ML domain. For example, in \cite{nassif2021machine}, common ML classifiers, such as KNN and random forest, have proved to be very effective in a more dynamic setting. Moving forward, as communication between IoTs and servers becomes more frequent, it is essential to have a more generic detector, which has the capability to not only detect based on past patterns, but also evolve as the communication packets become complicated. We believe the AGI has such capability.


\subsection{Security and Privacy}

The resource-constrained, distributed, and heterogeneous nature of IoT not
only impedes the deployment of AI algorithms but also entails acute
security and privacy issues that were never seen before. This section first
provides insights about how to safeguard AI workload on IoT devices at system
level. Then, we discuss algorithm-level security and privacy concerns. Finally,
we elaborate on the legitimate use of AGI in addressing IoT security/privacy problems.

\subsubsection{Safeguarding AI Workload on IoT} IoT devices generally run on
 less powerful hardware and lack many modern security mitigation mechanisms,
 making them more vulnerable than traditional computing platforms~\cite
 {yu2022building}. Worse, some IoT devices are deployed in ambient or
 unattended environments. This opens a door for adversaries to launch more
 powerful physical attacks such as cold-boot attacks to directly dump device
 memory~\cite{halderman2009lest,won2020practical}, leading to the theft of
 user privacy or manipulation of machine learning models. We envision three
 technological paths to mitigate these threats. 1) \textbf{Bug elimination:}
 Cyber attacks leverage firmware vulnerabilities to cause unintended
 behaviors. Therefore, extensive in-house firmware testing becomes essential
 to reduce cyber exploitation. A promising technique, known as firmware
 rehosting, tries to build a general IoT hardware model and emulate firmware
 execution on a virtualized hardware~\cite{P2IM,uemu}, making it possible to
 test thousands of instances simultaneously. 
 2) \textbf{Attack Mitigation:} Firmware testing can never eliminate
 all bugs. Therefore, firmware hardening can serve as the second line of
 defense to detect and thwart ongoing attacks. The idea is to instrument the
 firmware so that violations of predefined properties (e.g.,~control flow
 integrity and memory safety) can be actively detected. Firmware hardening
 often requires injecting additional code into the firmware, which inevitably
 imposes performance penalties. However, recent research has been successful
 in enforcing certain types of security properties (e.g.,~control flow
 integrity in $\mu$RAI~\cite{almakhdhub2020murai}) on resource-constrained
 embedded systems. 3) \textbf{Isolation:} Nowadays, our IoT
 firmware is bloated with libraries from multiple third-party contributors.
 It becomes necessary to isolate the confidential ML models from the rest of
 the firmware. By placing ML models in the hardware-enforced trusted
 execution environment (TEE), even if the firmware is compromised, the models
 cannot be accessed. Representative TEE implementations in IoT include ARM
 TrustZone-M~\cite{cortexmtrustzone} and RISC-V MultiZone~\cite
 {multizone}. To further reduce the code base in TEE, partitioned execution
 of ML only puts a security-critical portion in TEE~\cite
 {mo2022sok}. However, automatically splitting an ML model into a secure part
 and a non-secure part is a challenging task that warrants further research.

\subsubsection{Exploiting AGI for Bad}

Assuming a bug-free implementation, the ML algorithms themselves might be
vulnerable. Existing attacks on ML models can be largely applied to IoT AGI.
For example, in adversarial machine learning attacks, attackers find small
variations in inputs that can result in very different model outputs~\cite
{wallace2019universal}. In data poisoning attacks, attackers bias or
``poison'' the training data to compromise the resulting machine learning
models~\cite{sun2021data}. In model stealing attacks, attackers with
black-box access to the machine-learning-as-a-service systems aim to
duplicate the functionality of the model by stealing model's parameters~\cite
{tramer2016stealing,wang2018stealing}. Finally, denial-of-service attacks
disrupt the model’s availability by deliberately sending it high-cost
problems, aiming to overwhelm the host's resources to handle the inquiry.

With the emergence of foundation models in IoT AGI, a single point of failure
exacerbates the impacts of these traditional attacks~\cite
{bommasani2021opportunities}. Moreover, we found that the open nature of IoT
leads to new attack surfaces and brings about unprecedented challenges to the
ecosystem. In particular, the data collected on Internet-scale billions of
IoT devices makes it particularly challenging to validate the data integrity,
leading to data poisoning where falsified data is injected into the training
data. Such attacks can be stealthy and persistent in the sense that the
adversaries do not need to cause immediate damage. Rather, they inject a tiny
bit of falsified data each day and remain undetected for an extended period.

In smart homes, AGI can be used in home automation for personalized living
experiences. However, by learning home owners' looks and voices, the AI
technology can generate ``deep-fake'' audio and/or animated images, leading
to home robbery, unauthorized operations, etc. Using these biometric data,
social engineering attacks are possible by impersonating trusted individuals
or entities. Likewise, by embedding AGI into smart factories and self-driving
cars, an attacker could comprise AGI algorithms to cause factory equipment
shutdown and car accidents. In prompt injection attacks, a prompt is used to
make the model ignore previous instructions or perform unintended actions.
When running these AGI algorithms on IoT devices or the edge, the ML models
can be manipulated to generate false responses, leading to incorrect
decisions and actions. A prompt can also carry confidential data or source
code of a company, opening the door for challenges to compliance obligations
and putting intellectual property at risk.

 In training data extraction attacks, attackers recover raw training inputs
 that are memorized as a part of the model, leading to privacy issues~\cite
 {carlini2021extracting}. The data provider may not trust the training module
 in the first place. This is particularly concerning in handling IoT
 data (e.g.,~smart homes) where users' privacy is included. It has shown that
 GPT-4 achieves high accuracy and reliability in masking private information
 from unstructured medical texts~\cite{liu2023deid}. The same technique
 can be applied to redact sensitive user privacy from IoT data before they
 are presented to the AGI training module.

\subsubsection{Using AGI for Good} On the other hand, defenders can also
 weaponize themselves with such powerful tools to handle complex real-life
 IoT security problems at scale. For example, leveraging AGI, IT specialists
 can deploy systems to analyze device configurations and firmware versions,
 thus making recommendations for remediation (e.g., change insecure
 configurations or update vulnerable firmware). By analyzing system logs or
 network traffic, AGI can augment existing intrusion detection systems with
 its powerful NLP capability. In particular, IoT applications often come with
 natural language descriptions of their behaviors and data usage. By
 comparing the system logs captured at run-time and the claims extracted from
 the APP description, compliance checks can be conducted to detect
 violations. Once detected, AGI can further assist security analysts in
 analyzing and diagnosing the security threats, leading to faster discovery,
 response, and sharing.





\subsection{Persistent Challenges Beyond Current Solutions}
The challenge of IoT data storage lies in effectively managing and storing the vast and continuous stream of data generated by a multitude of interconnected devices and sensors \cite{abu2013data}. This data often comes in various formats and includes real-time updates, requiring robust storage solutions that can handle the volume, variety, velocity, and variability of the data. As IoT deployments expand. IoT ecosystems encompass an ever-growing number of devices and sensors, leading to an exponential increase in data volume. Traditional storage systems may struggle to handle such scale efficiently. IoT data can come in diverse forms \cite{krishnamurthi2020overview}, including structured, semi-structured, and unstructured data. Storage solutions must be flexible enough to accommodate various data formats and adapt to changes in data structure. It is challenging for low-cost sensors to accommodate the data variety and solubility, limiting the deployment scale of IoT systems. 

There are also challenges for local computing in IoT devices that need to be addressed for effective implementation and operation \cite{sehgal2012management}. Many IoT devices have constrained processing power, memory, and storage capacities. Running complex computations locally while ensuring efficient resource utilization can be a significant challenge. IoT devices are often battery-powered or have limited access to power sources. Local computing tasks must be optimized to minimize energy consumption and extend device battery life \cite{cui2018joint}. Some IoT applications require real-time processing of data for timely decision-making. Ensuring that local computations meet the required response times without introducing latency is crucial. Therefore, the complexities of algorithms in local computing would be limited, therefore potentially restricting the computing accuracy. Nevertheless, it is expected that with improved algorithm optimization efficiency and more advanced power-supplying devices, the challenges in data storage and computing can be effectively mitigated or resolved.  

In the realm of security and privacy, one of the pressing concerns is data encryption. As IoT devices proliferate, creating secure methods to encrypt data across a network becomes increasingly complex \cite{atlam2020iot}. Traditional encryption techniques may not be applicable, given that many IoT devices have resource constraints. Device authentication further compounds this issue. In a sprawling IoT network, establishing trust between various devices and central servers is complex and critical. A single unauthenticated or compromised device can become a gateway for network-wide vulnerabilities. Alongside this is the question of user privacy. IoT devices often collect and transmit personal information, from health data to geolocation. How this sensitive data is stored, accessed, and utilized is a subject that demands careful consideration, both from a technical and ethical standpoint \cite{kraijak2015survey}.

Scalability and interoperability pose another set of challenges \cite{rana2021internet}. With IoT devices coming from a plethora of manufacturers, each with its own set of standards and protocols, achieving seamless interoperability is a daunting task. This lack of standardization is not merely a technical obstacle but a barrier to scaling IoT networks to accommodate the exponential increase in devices. In the absence of universal protocols, interoperability, and scalable architectures become significant hurdles, inhibiting the efficient growth and functionality of IoT ecosystems.

Environmental and social challenges are often overlooked but are no less significant \cite{matta2019internet}. While IoT devices offer potential solutions for environmental monitoring, they themselves can be quite energy-intensive, thus contributing to the very problems they aim to solve. This irony underscores the need for designing more energy-efficient devices. The rapid development and turnover of IoT technology also lead to another concern: electronic waste \cite{dauvergne2020ai}. The limited lifespan of these devices exacerbates existing environmental issues related to e-waste, a challenge yet to be fully tackled.

Regulation and governance in IoT are largely nascent and nebulous \cite{pagallo2017new}. Despite the rapid technological advancements, the legal framework governing IoT remains ambiguous, creating challenges for accountability and compliance \cite{rosner2018clearly}. This is particularly critical when IoT technologies are implemented in sectors with stringent regulations, such as healthcare \cite{liu2023deid} and transportation. The absence of clear regulatory guidance can lead to ethical and legal grey areas, putting both users and providers in precarious positions.

Last but not least, human factors should not be underestimated. The usability of IoT devices can vary widely depending on the design, and the target user demographic \cite{zheng2018user}. Crafting interfaces that are both intuitive and functional across different devices and user groups is a complex challenge \cite{liu2020survey,bancilhon2020let,bancilhon2019let,nathoo2020using}. Moreover, ethical concerns about how data collected by IoT devices is used can arise, including issues related to surveillance or discrimination based on data analytics \cite{wachter2018normative}. Ensuring ethical use while maintaining utility is a challenge that has social, cultural, and political implications.

Addressing these challenges in IoT will require a multidisciplinary approach, involving contributions from various fields including cybersecurity, legal studies, environmental science, and social sciences. While these challenges present significant obstacles, they also open avenues for innovation and advancements in technology and policy.


\section{Conclusion}
In this survey, we discuss how Artificial General Intelligence (AGI) can revolutionize the Internet-of-Things (IoT) systems and frameworks, and at the same time introduce unique challenges. AGI can be used to apply knowledge gained in one domain in the heterogeneous data collected in IoT systems to new domains, integrate and contextualize data from diverse sensory inputs, understand causes and effects in events, discover, communicate, and interact with humans on dynamic and complex systems, and learn and apply new skills to unseen tasks. AGI-powered IoT systems have a wide range of applications including smart grids, home, agriculture, manufacturing, healthcare, and transportation. In the meantime, those applications require significant new research in AGI to be adapted to resource-limited scenarios in IoT. Data privacy and security also need to be considered in designing AGI-powered IoT systems.

While this article summarizes the diverse opportunities and applications enabled by AGI, we expect to see new and exciting applications powered by AGI in IoT with the rapid evolution of technology in AI and machine learning. We hope this survey can inspire researchers from both the AI as well as the IoT communities to develop and explore the limitless possibilities of AGI+IoT.

\ifCLASSOPTIONcaptionsoff
  \newpage
\fi



%
\bibliographystyle{ieeetr}
\bibliography{ref.bib}




%






\end{document}